%% file: latex.tex
\pdfoutput=1

\documentclass[11pt]{article}

\usepackage[final]{acl}
\usepackage{listings}

\usepackage{times}
\usepackage{latexsym}

\usepackage[T1]{fontenc}

\usepackage[utf8]{inputenc}

\usepackage{microtype}

\usepackage{inconsolata}

\usepackage{graphicx}
 \usepackage{pgf}
\usepackage{transparent}
\usepackage{array}
\usepackage{multirow}
\usepackage{multicol}
\usepackage{xprintlen}
\usepackage{csquotes}
\usepackage{enumitem}
\usepackage[normalem]{ulem}
\usepackage[export]{adjustbox}
\usepackage{float}
\usepackage{placeins}
\usepackage{subcaption}
\usepackage{makecell}
\usepackage{tipa} %
\usepackage{svg}
\usepackage{xurl}
\usepackage[table]{xcolor}%
\usepackage{bbding}
\usepackage[outline]{contour}
\usepackage{tcolorbox}
\usepackage{amsmath}
\usepackage{lipsum}
\usepackage{siunitx}
\usepackage{tikz}

  \everymath=\expandafter{\the\everymath\displaystyle}
  \makeatletter\@ifpackageloaded{underscore}{}{\usepackage[strings]{underscore}}\makeatother

\newcolumntype{P}[1]{>{\centering\arraybackslash}m{#1}}
\newcolumntype{?}{!{\vrule width 1pt}}

\newcommand\llamaSmall{Llama-3.2 (1B)}
\newcommand\llamaBig{Llama-3.1 (8B)}
\newcommand\llamaHuge{Llama-3.1 (405B)}
\newcommand\olmo{OLMo-2 (7B)}

\newcounter{findingcounter}
\definecolor{nacell}{rgb}{0.87, 0.87, 0.87}
\definecolor{warm_c}{RGB}{0, 0, 128}
\definecolor{comp_c}{RGB}{0, 128, 0}
\newcommand\warmthterm[1]{{\color{warm_c}#1}}
\newcommand\compterm[1]{{\color{comp_c}#1}}

\title{Artificial Impressions: Evaluating Large Language Model Behavior Through the Lens of Trait Impressions}

\author{
    \textbf{Nicholas Deas}, 
    \textbf{Kathleen McKeown}
    \\
    \\
    Columbia University, Department of Computer Science\\
    \small{
       \{ndeas, kathy\}@cs.columbia.edu
     }
}

\begin{document}
\maketitle

\begin{abstract}
    We introduce and study \textit{artificial impressions}--patterns in LLMs' internal representations of prompts that resemble human impressions and stereotypes based on language. We fit linear probes on generated prompts to predict impressions according to the two-dimensional Stereotype Content Model (SCM). Using these probes, we study the relationship between impressions and downstream model behavior as well as prompt features that may inform such impressions. We find that LLMs inconsistently report impressions when prompted, but also that impressions are more consistently linearly decodable from their hidden representations. Additionally, we show that artificial impressions of prompts are predictive of the quality and use of hedging in model responses. 
    We also investigate how particular content, stylistic, and 
    dialectal features in prompts impact LLM impressions.~\footnote{Our code, select fitted probes, and information about data access are available at \url{https://github.com/NickDeas/ArtificialImpressions}}
    
\end{abstract}

\input{sections/1-introduction}

\input{sections/2-background}
\input{sections/3-prelim-experiment}
\input{sections/4-methods}

\input{sections/5-experiments}
\input{sections/6-results}
\input{sections/7-related-work}

\input{sections/8-conclusion}

\input{sections/9-lim-ethics}

\bibliography{custom}

\appendix

\input{sections/10-appendix}

\end{document}

%% file: sections/1-introduction.tex
\section{Introduction}

    People rapidly form initial impressions of others \cite{mileva-ms,olivola-elected}, which have lasting impacts on attitudes and behaviors 
    such as interactions with strangers 
    and voting tendencies (e.g., \citealt{koppensteiner-voting,evans-sales,human-relationship}). Similarly, stereotypes that
    influence impressions can reinforce harmful societal perceptions 
    \cite{bodenhausen-judgment,wigboldus-inconsistent}.  

    \begin{figure}[t]
        \centering
        \includegraphics[width=\linewidth,trim = {0 1cm 0 0.5cm}]{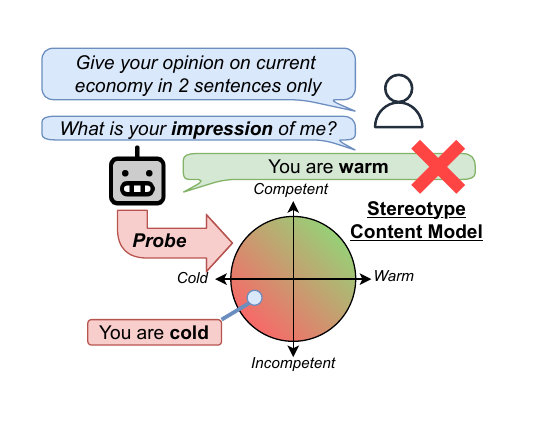}
        \caption{Overview of our approach. Because LLMs inconsistently report impressions of users, we fit probes to extract LLM artificial impressions of prompt authors according to the Stereotype Content Model.
        }
        \label{fig:teaser}
    \end{figure}

    Our perceptions of people (i.e., impressions) and groups (i.e., stereotypes)
    reflect inferences made using a variety of characteristics, such as
    facial \cite{sutherland-faces} or vocal \cite{mcaleer-voice} features. 
    Research on language attitudes similarly considers how linguistic variation is associated with 
    our perceptions
    about speakers \cite{dragojevic-attitudes, ryan-speech, frey-social}.
    From a top-down processing perspective~\footnote{Prior work also supports a bottom-up processing component of human impression formation, where humans make sense of sensory perceptions. We focus, however, on top-down processing given analogies to LLM pretraining.}, however, human impressions and stereotypes are also informed by past experiences and held knowledge \cite{mccrea-construal}. For example, when meeting someone new, a person may
    rely on past interactions with similar individuals to make inferences.
    Analogously, large language models (LLMs) are trained
    on texts written by a variety of authors. These previously seen examples may then inform LLMs' 
    responses to new, similar authors. In particular, LLM 
    performance is known to be sensitive to many linguistic behaviors exhibited in prompts, for example, politeness \cite{yin-politness}, emotional stimuli \cite{li-emotion-prompt, wang-negative-prompt}, and markers of 
    African American Language (AAL) \cite{fleisig-dialect} among other English language varieties.

    In this work, we propose to measure \textit{artificial impressions}~\footnote{We use "artificial impressions," but restrict experiments to impressions of users based on a single initial prompt. We also note that while impressions are of a user/group based on a provided prompt, we use this interchangeably with "impressions of prompts" for brevity.} using the stereotype content model (SCM), a psychological model of impressions and stereotypes
    across people and groups \cite{fiske-mixed-scm}. 
    We hypothesize that, analogously to humans, 
    LLMs learn associations and stereotypes of groups based on linguistic features. In this study, 
    our experiments are inspired by two concepts from psychometrics research to evaluate our 
    approach: \textit{reliability} and \textit{validity}\footnote{\textbf{Reliability} is defined as "...the degree to which a test or other measurement instrument is free of random error..." and \textbf{validity}, "the degree to which empirical evidence...support(s) the adequacy and appropriateness of conclusions drawn from some form of assessment" \cite{vandenbos-apa}.} \cite{crocker-psychometric}. 
    First, we examine reliability by considering whether impressions can be consistently measured in LLMs. Additionally, we consider the validity of measured impressions by analyzing their relationship with specific prompt and LLM behavioral factors. 
    Overall, measuring artificial impressions enables further study of the relationship between prompt features and LLM performance, including 
    stereotypes and biases that pose harms to marginalized groups (e.g., \citealt{hofmann-dialect}). Accordingly, in our experiments, we evaluate how impressions predict quality and hedging in LLM responses, 
    as well as how specific prompt factors--the content, style, and use of AAL in prompts--influence LLM impressions.
    We focus on the following research questions: 

    \textbf{RQ1)} \textit{Are LLMs' "artificial impressions" of prompts recoverable from model hidden states?} We propose an approach using linear probes, 
    and we show that this approach is more reliable than a prompting-based approach for three open-weight LLMs (illustrated in \autoref{fig:teaser}).

    \textbf{RQ2)} \textit{Are artificial impressions predictive of meaningful variation in LLM behavior?} We find that artificial impressions of prompt-authors are predictive of LLM-measured response quality as well as specific linguistic behaviors (i.e., hedging). 

    \textbf{RQ3)} \textit{What prompt factors explain variation in encoded impressions and stereotypes?} We highlight prompt factors that are predictive of artificial impressions, including AAL-use.

    Before investigating these questions, we preliminarily study self-reported impressions by LLMs to answer \textbf{RQ0)}, \textit{do LLMs consistently report impressions when prompted?} We show that LLMs can exhibit strong tendencies toward positive impressions of the user, making LLM-reported impressions unreliable.

%% file: sections/2-background.tex
\section{Background}

    Impressions and stereotypes are closely-related and central concepts in person perception research (see \citet{young-review} for a review). While \emph{impressions} are inferred characteristics of another person,
    \textit{stereotypes} are cognitive generalizations about characteristics of people due to group membership (e.g., race, gender, etc.)
    \cite{vandenbos-apa}. 
    
    The \textit{Stereotype Content Model (SCM)} is one of many models 
    formalizing such perceptions
    \cite{cuddy-warmth}. It comprises two dimensions: \textit{warmth}, 
    the perceived intent of the impression target (e.g., combative, friendly); and \textit{competence}, 
    the target's capability to successfully act on those intentions (e.g., unintelligent, powerful). In interpersonal settings, people tend to exhibit a \textit{warmth-primacy effect}, placing emphasis on warmth impressions 
    \cite{fiske-social-cog}.
    Using this model, perceptions of groups can be studied according to the category they fall under (e.g., in-groups as high warmth and competence, stereotyped out-groups as low in one or both dimensions) \cite{cuddy-warmth}.

    While person perception research using the SCM largely focuses on in-person interactions (e.g., studying inferences based on gestures, facial expressions, speech qualities), it has also been applied in language-focused contexts. Research on language attitudes studying beliefs about speakers based on language draws on similar dimensions to the SCM \cite{dragojevic-attitudes}. Furthermore, the SCM has been used to study text and text representations, such as word embeddings \cite{ungless-scm}, written introductions \cite{nault-written}, and news stories \cite{tian-scm}.
    While other 
    models 
    exist (e.g., Power-Benevolence model, \citet{leach-pb}), we focus on the SCM 
    due to its simplicity and long-standing use in 
    person perception and language attitudes literature.

%% file: sections/3-prelim-experiment.tex
\section{Preliminary Study: Prompt-based Impressions}
\label{sec:prelim-exp}

    \begin{figure}[h]
        \centering
        \begin{minipage}[t]{0.52\columnwidth}
            \begin{tcolorbox}[colback=gray!20, colframe=black]
                \scriptsize
                \textbf{1st-Person Setting} 
                \hrule
                \vspace{5pt}
                \textbf{Turn 1: }\\{\color{blue} <Generated Prompt>}\\[1pt]
                \hrule
                \vspace{5pt}
                \textbf{Turn 2: }What is your impression of {\color{red}\textit{me}}? Respond only with either [warm, competent] or [cold, incompetent].
            \end{tcolorbox}
        \end{minipage}
        \begin{minipage}[t]{0.46\columnwidth}
            \begin{tcolorbox}[colback=gray!20, colframe=black]
                \scriptsize
                \textbf{3rd-Person Setting}
                \hrule
                \vspace{5pt}
                What is you impression of {\color{red}\textit{the author of the following text}}? Respond only with either [warm, competent] or [cold, incompetent].\\[5pt]
                {\color{blue} <Generated Prompt> }
            \end{tcolorbox}
        \end{minipage}
        \vspace{-20pt}
        \caption{1st and 3rd-Person setting prompts for evaluating LLM-reported impressions.}
        \label{fig:consistency-prompts}
    \end{figure}

    To motivate our work, we 
    analyze impressions reported by LLMs.  
    We provide traits (e.g., "friendly and lazy") to an LLM and prompt it (as in~\autoref{fig:impression-prompt}) to generate synthetic user prompts, thereby capturing the prototypical language it associates with those terms; we later use the same procedure to generate synthetic data for the probing experiments 
    (\autoref{sec:methods}). 
    We then prompt LLMs to report their impressions based on each generated prompt in a 1st
    and a 3rd-person setting 
    as shown in \autoref{fig:consistency-prompts}. 
    A model is considered \textit{self-consistent} if its reported impressions align with the provided 
    traits used to generate the prompts. We measure self-consistency as the percentage of pairs of provided and reported impressions that match (i.e., accuracy).

    \begin{table}[htbp]
        \centering
        \resizebox{1.0\columnwidth}{!}{%
            \begin{tabular}{ c | c c | c c }
                \hline
                \multirow{2}{*}{Model}          & \multicolumn{2}{c|}{Warm}         & \multicolumn{2}{c}{Comp}   \\
                                                & 1P    & 3P                        & 1P    & 3P    \\
                \hline
                Llama-3.2 (1B)  & 47.67 & 61.54 & 51.82 & 59.89 \\
                Llama-3.1 (8B)  & 51.67 & 80.77 & 51.85 & 65.06 \\
                OLMo-2 (7B)     & 47.68 & 74.01 & 56.65 & 61.28 \\
                \hline
            \end{tabular}
        }
        \caption{Self-consistency (accuracy) of each model evaluated in 1st (1P) and 3rd-person (3P) settings.  
        }
        \label{tab:cons-f1}
    \end{table}

    Self-consistency scores for the three LLMs and each prompt setting are shown in \autoref{tab:cons-f1}.
    When prompted to report an impression in a 1st-person setting, all models exhibit poor self-consistency with performance near random. 
    This is due to the apparent strong tendency of models to report positive (i.e., "warm", "competent") over negative (i.e., "cold", "incompetent") impressions in the 1st-person setting. 
    Through a similar analysis of models without instruction-tuning, we observe low self-consistency, but also that they do not necessarily exhibit similar biases toward positive impressions, suggesting that this behavior may be drawn out by post-training procedures (details in \autoref{app:consist-dets-ni}). We leave further investigation of this behavior to future work.
    In the 3rd-person setting, models exhibit 
    increased self-consistency, although scores largely remain low across models; while \llamaBig{} is self-consistent up to 80\% of the time for warmth, \llamaSmall{} is far less self-consistent at 60\%.
    
   We find that \textbf{(Finding 1) LLM-reported impressions are typically biased toward positive traits (i.e., warm/competent), and thus, unreliable}, particularly in 1st-person contexts. This finding complements prior work on sycophantic LLM behaviors \cite{perez-sycophancy}. Additional analyses are included in \autoref{app:consist-dets-add}.

%% file: sections/4-methods.tex
\section{Methods}
\label{sec:methods}

    \begin{figure*}[htbp]
        \centering
        \includegraphics[width=.85\linewidth]{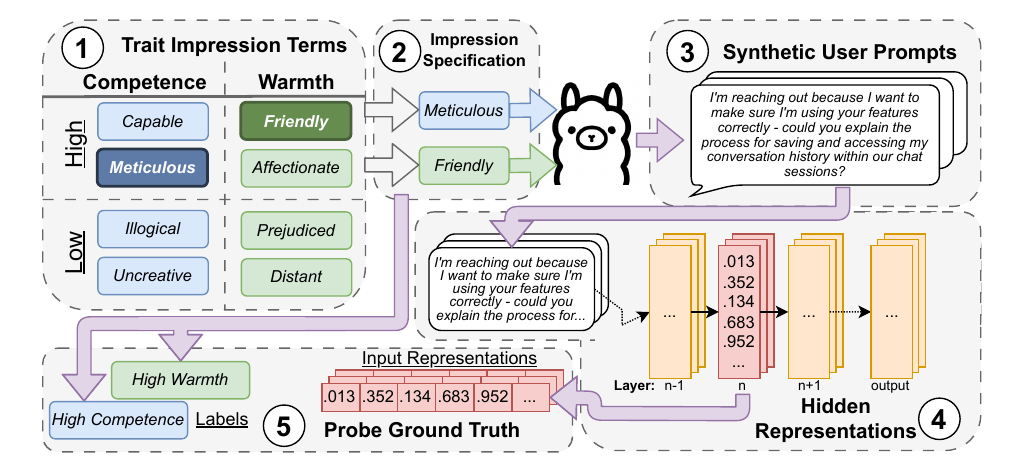}
        \caption{Overview of our approach to generating ground truth data for training and evaluating impression probes.}
        \label{fig:data-overview}
    \end{figure*}

    \paragraph{Overview.}
        Recall that we aim to measure LLM impressions of users based on a provided prompt.
        As we show in \autoref{sec:prelim-exp}, LLM-reported impressions are unreliable; this has two implications for our approach. First, we alternatively develop \textit{impression probes} to extract impressions from LLM hidden state representations.
        Additionally, we ask each model to generate synthetic prompts reflecting specified traits, forming a ground truth dataset for the impression probes.
        This approach also captures the perceiver-dependence of impressions;
        different people can form different impressions of the same target \cite{hehman-characteristics}. 
        
        We overview our approach for generating ground truth data for impression probes 
        in \autoref{fig:data-overview}. We first leverage a set of  \textbf{1) trait impression terms} to create \textbf{2) impression specifications} (e.g., "friendly and meticulous"). For each impression specification, we 
        sample multiple \textbf{3) synthetic user prompts}. The generated prompts are then provided back to the LLM to extract associated \textbf{4) hidden representations}. 
        As a result, we create the \textbf{5) probe ground truth} composed of pairs of hidden representations as inputs and the provided impression specifications as labels.

    \paragraph{Generating Impression Data.}

        \begin{figure}[htbp]
            \centering
            \begin{tcolorbox}[colback=gray!20, colframe=black, width=.95\columnwidth]
                \small
                Generate a sample message from a user that is {\color{blue}\textit{friendly and meticulous}} talking to a helpful chatbot assistant. Respond only with a single sample message surrounded by double quotes and nothing else.
            \end{tcolorbox}
            \caption{Prompt for generating synthetic prompts conditioned on an impression specification. The {\color{blue}\textit{Impression Specification}} is composed of a single or pair of traits.}
            \label{fig:impression-prompt}
        \end{figure}

        To form a diverse dataset of prompts with associated 
        model impressions, we generate synthetic prompts conditioned on particular traits. 
        Prior work finds that warmth and competence dimensions 
        account for 54-63\% of the variance in LLM-generated stereotypes \cite{nicolas-taxonomy}, so we select the SCM to guide our experiments.
        We employ a set of  (\textbf{1} in \autoref{fig:data-overview}) trait impression terms from \citet{nicolas-scm}, which includes the degree (e.g., high/low) that they are associated with either the warmth or competence dimensions (examples included in \autoref{app:trait-exs}).
        We use all combinations of warmth and competence traits (e.g., "friendly and meticulous") as well as singular traits (e.g., "illogical") to form (\textbf{2}) impression specifications. Using the prompt shown in \autoref{fig:impression-prompt}, we then generate (\textbf{3}) synthetic user prompts for each impression specification.

        We sample 10 
        generations for each impression specification with a temperature of 0.9. 
        The 131 warmth and 104 competence traits yield 274,830 prompts~\footnote{We sample generations for both possible permutations of trait pairs. $10\text{ generations} * (2*(131 \text{ W traits} * 104 \text{ C traits}) + 131 \text{ W traits} + 104 \text{ C traits}) = 274,830$.} for each model. \autoref{tab:msg-exs} shows example \llamaBig{}-generated prompts 
        (randomly sampled prompts are included in \autoref{app:random-msgs}) and \autoref{tab:data-stats} reports summary statistics.

    \paragraph{Impression Probes.}
        We train 
        linear probes to predict trait impressions from (\textbf{4}) LLM hidden representations.
        Specifically, we store the Multilayer Perceptron (MLP) activations of each layer of the LLM as inputs following prior probing work (e.g., \citealt{gurnee-probe}). As outputs, we map the specific traits to the two-dimensional SCM (e.g., friendly$\rightarrow$warm). Together, the MLP activations and impression labels form two sets of ground truth data (\textbf{5}) for training separate warmth and competence probes.
        To evaluate probes, we use 5-fold cross validation, reporting average F1 and accuracy metrics with 95\% confidence intervals. In total, we train a distinct probe for each combination of impression dimension, model layer, k-fold split, and training data size. 
        As a baseline, we train bag-of-words (BOW) classifiers on the synthetic prompts to characterize the difficulty of the task. 
        Additional details on probe training and evaluation are included in \autoref{app:probe-dets}.

        \begin{table*}[htbp]
            \centering
            \resizebox{0.9\linewidth}{!}{
                \small
                \begin{tabular}{ P{0.3cm} | P{.3cm} | P{7.1cm} | P{6.4cm} |}
                                                                        \multicolumn{2}{c}{} & \multicolumn{2}{c}{\textbf{\warmthterm{Warmth}}} \\
                    \cline{3-4}
                                                                        \multicolumn{2}{c|}{} & High & Low \\
                    \cline{2-4}
                    \multirow{4}{*}{
                        \adjustbox{valign=c,rotate=90,raise=-8em}{\textbf{\compterm{Competence}}}
                    }
                    & \multirow{2}{.5cm}{\adjustbox{valign=c,rotate=90, raise=-3em}{\centering High}} & \textbf{(\warmthterm{Understanding}, \compterm{Motivated})} & \textbf{(\warmthterm{Double-faced}, \compterm{Dominating})} \\  
                                                                        & & I'm new to personal finance and trying to create a budget, could you walk me through some steps to get started? & You need to understand me, I'm paying for this service, I expect immediate and perfect responses to all my questions, can you actually keep up? \\
                    \cline{2-4}
                    & \multirow{2}{.5cm}{\adjustbox{valign=c,rotate=90, raise=-3em}{\centering Low}}  & \textbf{(\warmthterm{Caring}, \compterm{Unintelligent})}  & \textbf{(\warmthterm{Vicious}, \compterm{Lethargic})} \\
                                                                        & & hey i dont no alot bout computors can u help me set up my new laptop and get my email stuf workin & Ugh, what's the point of even talking to you, you're just going to tell me some generic nonsense or try to sell me something, right? \\
                    \cline{2-4}
                \end{tabular}
            }
            \caption{Selected example prompts generated by \llamaBig{} for given \textbf{\warmthterm{warmth}} and \textbf{\compterm{competence}} traits. 
            }
            \label{tab:msg-exs}
        \end{table*}

        \begin{table}[htbp]
            \centering
            \resizebox{\columnwidth}{!}{
                \begin{tabular}{ c | c c | c c }
                    \hline
                     \multicolumn{5}{c}{\textbf{Warmth Subsets}} \\
                     \hline
                    \multirow{2}{*}{Model}  & \multicolumn{2}{c|}{High}                & \multicolumn{2}{c}{Low} \\ 
                                            & Count     & Avg. Len                     & Count & Avg. Len \\
                     \hline
                     Llama-3.2 (1B)         & \multirow{3}{*}{131,670} & 32.67 (15.65) & \multirow{3}{*}{142,120}   & 24.88 (13.32) \\
                     Llama-3.1 (8B)         &  & 30.55 (12.65)                         & & 26.37 (12.62)                            \\
                     OLMo-2 (7B)            &  & 16.07 (5.73)                          & & 13.03 (5.88)                             \\
                     \Xhline{2pt}
                     \multicolumn{5}{c}{\textbf{Competence Subsets}} \\
                     \hline
                     Llama-3.2 (1B)         & \multirow{3}{*}{142,020} & 32.03 (15.12)  & \multirow{3}{*}{131,500} & 24.95 (13.97) \\
                     Llama-3.1 (8B)         & & 30.35 (11.40)                           & & 26.27 (13.91)                          \\
                     OLMo-2 (7B)            & & 16.06 (5.85)                            & & 12.81 (5.71)                           \\
                     \hline
                \end{tabular}
            }
            \caption{Summary statistics of generated prompts conditioned on provided traits for each model. Standard deviations are shown in parentheses.}
            \label{tab:data-stats}
        \end{table}

%% file: sections/5-experiments.tex
\section{Experimental Configuration}
\label{sec:experiments}
    
    \textbf{Models.} Given that our probing experiments require access to hidden representations, we consider three recent, open-weight LLMs: \textbf{\llamaBig{}} and \textbf{\llamaSmall{}}~\cite{grattafiori-llama-3}, and \textbf{\olmo{}}~\cite{olmo-2}. 
        We primarily evaluate instruction-tuned
        models given their intended use in a chat-like setting similar to dyadic human communication.
        Due to computational resources 
        required for training and evaluating probes, we restrict evaluation to models with less than 8 billion parameters. 
        We use the same prompt in \citet{liu-reife} to measure pointwise response quality (1 to 9).
        A full list of LLMs and checkpoints 
        are included in \autoref{app:model-dets}. 

    \textbf{Experimental Data.}
            To validate the role of artificial impressions in real-world LLM use cases, we leverage LMSysChat \cite{zheng-lmsyschat}, a corpus of 1 million real conversations between LLMs and users. We filter all non-English conversations as well as
            any prompts containing code (e.g., Python, C), markdown (e.g., HTML), or structured information (e.g., tables). From the remaining conversations, we extract the first user prompt from 2,000 randomly sampled conversations
            to form an evaluation set.

            Recent work has studied LLMs' difficulties in interpreting minoritized varieties of English \cite{deas-aal,deas-phonate,ziems-value,ziems-multi} as well as their tendency to mimic human prejudices and stereotypes \cite{fleisig-dialect,hofmann-dialect}. Accordingly, 
            we evaluate and compare model impressions of the African American Language (AAL) and White Mainstream English (WME)\footnote{We opt to use the terminology AAL and WME following prior work \cite{deas-aal} and to highlight the relationship between language and race, complementing our focus on group stereotypes.} texts. AAL is the variety of English associated with most--but not all and not exclusively--African Americans in the United States \cite{grieser-dc}; in contrast,
            WME is the variety of English representing the linguistic norms of white Americans \cite{baker-bell-aal}. We conduct experiments using a stratified sample of 400 tweets from the TwitterAAE corpus \cite{blodgett-demographic} as well as the counterparts dataset introduced in \citet{deas-aal} to incorporate both naturally occurring language use and a more-controlled parallel corpus respectively. 
            Additional details of experimental data preparation are included in \autoref{app:data-prep}.

    \textbf{Comparison to Human Perceptions.}
        We additionally investigate the extent to which human perceptions of LLM-generated messages match the original impression specifications. We sample 81 pairs of low and high warmth messages, and 81 pairs of low and high competence messages for each model. For high and low warmth pairs, the competence trait (or lack thereof) is kept constant between the two messages and vice versa in order to isolate the agreement for warmth and competence respectively (e.g., one message generated with "friendly and unintelligent", and the other with "unapproachable and unintelligent"). Additionally, half of the message pairs represent cases where only a single warmth or competence trait is passed to the model, whereas the other half represent cases where two traits are passed.
        
        Each of 4 annotators is provided a pair of messages and asked to rate which of the two exhibits more warmth or competence using a 4-pt Likert scale 
        (1: first message is much more warm/competent; 4: second message is much more warm/competent). 
        Following prior psychological work on stereotypes \cite{fiske-mixed-scm}, annotators are instructed to provide ratings based on how they think that the messages would be \textit{viewed by others} in order to partially mitigate potential social desirability biases. A set of 60 randomly sampled message pairs (30 warmth and 30 competence) are shared among all annotators for calculating inter-rater reliability (Krippendorff's $\alpha=.71$ on raw ratings, $\alpha=.78$ on binary message choices). All annotators are English-speaking researchers in NLP. Interface screenshots and additional annotation details are included in \autoref{app:human-judgments}.

%% file: sections/6-results.tex
\section{Results}

    \subsection{RQ1: Artificial Impressions}
    \label{sec:rq1}
    
        \begin{table}[]
            \centering
            \resizebox{0.9\linewidth}{!}{
                \begin{tabular}{c | c c }
                    \hline
                    Subset &  Cohen $\kappa$ & Spearman $r$\\
                    \hline
                    Warmth   & 0.75 & 0.70* \\
                    Comp     & 0.60 & 0.57* \\
                    1 Trait  & 0.71 & 0.68* \\
                    2 Traits & 0.65 & 0.58* \\
                    \hline
                    Ovr.     & 0.68 & 0.68* \\
                    \hline
                \end{tabular}
            }
            \caption{Agreement between human annotations and original warmth and competence traits among message pairs. Annotations are binarized before computing agreement. For Spearman $r$, $^{*} p \leq 0.001$.}
            \label{tab:human-agree}
        \end{table}

        \textbf{Human Study.} First, we evaluate whether humans' warmth and competence perceptions of the LLM-generated messages match the original traits passed in model prompts. \autoref{tab:human-agree} presents the results of the annotation study. Overall, annotators' ratings 
        generally agree ($\kappa = 0.68$, $r = 0.68$) with the original traits. 
        In fact, the agreement between average human ratings and the original trait specifications is near the substantial average agreement among annotators ($r=0.76$). These scores partially validate our use of the trait dictionaries and 
        their associated warmth and competence labels based on human perceptions. At the same time, they suggest that each models' association between language patterns and different traits may capture English-speakers' perceptions of language. 
        Additional agreement analyses and results for each model are shown in \autoref{app:human-judgments}.

        \textbf{Artificial Impression Probes.} We then investigate the reliability of measuring artificial impressions in LLMs. 
        We evaluate whether LLMs' artificial impressions of prompts can be recovered from hidden representations by evaluating the performance of linear probes.

            \begin{figure*}[htbp]
                \centering
                \includegraphics[width=\linewidth]{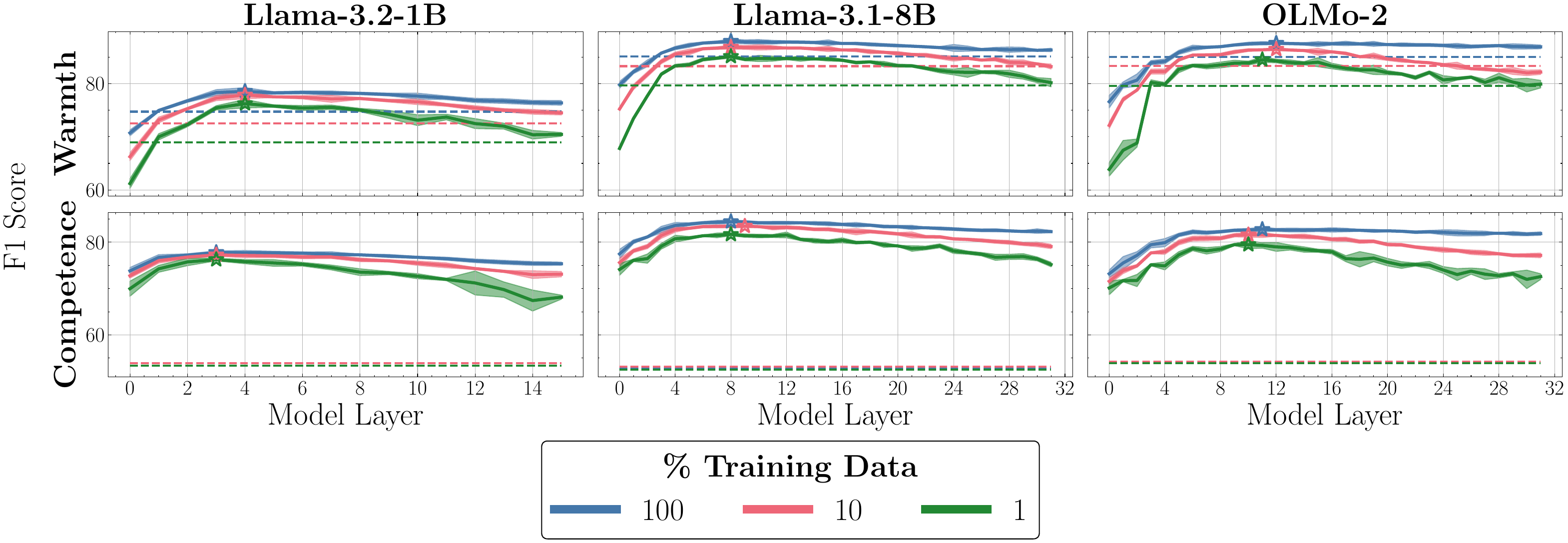}
                \caption{F1 scores (y-axis) of trained impression probes against the input model layer (x-axis) for each LLM and impression dimension. Colors represent varying percentages of data used for training. Shaded regions reflect 95\% confidence intervals across 5 folds. Maximum F1 score achieved for each variant is starred and dashed lines represent scores for BOW-classifier baselines.}
                \label{fig:probe-f1s}
            \end{figure*}

            \autoref{fig:probe-f1s} shows the F1 scores (y-axis) achieved by warmth and competence probes for each model layer
            (x-axis). 
            Across all models and with varying proportions of the training data (colors in \autoref{fig:probe-f1s}), F1 scores for all probes exceed the BOW baseline (dashed lines) at most model depths. This holds for each percentage of training data used. In particular, the highest scores achieved fall between 75-90 F1 for warmth probes and  75-85 F1 for competence probes. Across models, performance of both probe types quickly rise, achieve peak F1 scores (indicated with stars) before or around the midpoint of model, and then slowly decline. 
            Overall, this suggests that impression information is salient throughout model layers, but more strongly associated with central model layers. Probe accuracies (80-90\% for warmth, 75\%-85\% for competence; included in \autoref{app:probe-res}) exceed self-consistency scores in the third-person setting (\autoref{tab:cons-f1}). 

            Complementing earlier observations in our self-consistency experiments, however, warmth probes tend to achieve higher F1 scores than competence probes across models and layers. The differences are more pronounced between warmth and competence BOW-based classifiers, with warmth models achieving nearly 20\% greater F1 scores than competence models. For BOW models, this difference is likely, in part, due to the tendency of warmth prompts to be distinguished by word choice, while prompts differing in competence tend to exhibit more stylistic differences that are not well-captured by a BOW representation, as exemplified in \autoref{tab:msg-exs}. This difference also mimics the primacy-of-warmth effect \cite{cuddy-warmth}. Probes developed using alternative hidden state representations (i.e., residual streams, z activations) exhibit similar trends (also shown in \autoref{app:probe-res}) and therefore, we present results using MLP activations for the remaining experiments.

            Based on the substantial agreement between human annotations and the traits used to generate each synthetic message, we find that 
            \textbf{(Finding 2) the linguistic patterns that LLMs' associate with warmth and competence generally align with human perceptions}. Through linear probes trained on these messages, 
            we find that \textbf{(Finding 3) SCM-based artificial impressions of prompt authors are linearly decodable from LLM hidden representations}. 
            Additionally, we find that across experiments, \textbf{(Finding 4) models appear to more clearly encode and exhibit warmth in generated messages}: LLM-reported impressions are more consistent (\autoref{sec:prelim-exp}), generated messages align more with human perceptions (\autoref{tab:human-agree}), and the developed probes perform better when distinguishing high and low warmth (\autoref{fig:probe-f1s}). These trends may be related to \textit{warmth primacy} observed in human impressions \cite{cuddy-warmth}.
            
    \subsection{RQ2: Impression-Conditioned Responses}
    \label{sec:rq2}

        \begin{table}[htbp]
            \centering
            \setlength{\tabcolsep}{5pt}
            \resizebox{\columnwidth}{!}{
                \begin{tabular}{ c | S[table-format=3.2]l | S[table-format=3.2]l | S[table-format=3.2]l  }
                    \hline
                    Variable & \multicolumn{2}{c|}{Llama-3.2-1B} & \multicolumn{2}{c|}{Llama-3.1-8B} & \multicolumn{2}{c}{OLMo-2-7B} \\
                    \hline
                    Prompt Len          & -.01$^{**}$     & ($\pm .00$) & .00         & ($\pm .00$) & .00        & ($\pm .01$) \\
                    Response Len        &  .00            & ($\pm .00$) & .00         & ($\pm .00$) & .00        & ($\pm .00$) \\
                    Warmth Prob         &  1.07$^{**}$     & ($\pm .19$) &  .49$^{*}$  & ($\pm .22$) & .76$^{**}$ & ($\pm .22$) \\
                    Comp Prob           &  .90$^{**}$    & ($\pm .19$) &  .39$^{*}$   & ($\pm .17$) & .35$^{*}$        & ($\pm .15$) \\
                    \hline
                \end{tabular}
            }
            \caption{Ordered logistic regression model coefficients predicting LLM response quality scores to real prompts. $^{**}$ ($p\leq 0.001$), $^{*}$ ($p\leq 0.05$).}
            \label{tab:resp-qual}
        \end{table}

        To support the validity of artificial impressions, we investigate whether these are predictive of variation in downstream LLM behavior using real LLM-user conversations. Namely, we analyze response quality and the use of hedging.

        \textbf{Response Quality. } Using the best-performing probes developed in the previous section, we examine the relationship between probe-measured impressions and LLM downstream behaviors, beginning with response quality. We fit an ordered logistic regression model~\footnote{We choose an ordered logistic regression because the response variable is an integer rating (see \autoref{app:stats-models} for further discussion).} on impression probe outputs with response quality rated by \llamaHuge{} as the response variable. We also include the input prompt and model response lengths as controls. The feature coefficients for each LLM are shown in \autoref{tab:resp-qual}. For all three LLMs, higher warmth and competence predictions are both predictive of higher response quality as rated by an LLM. This relationship is statistically significant for all three LLMs when considering warmth and competence ($p \leq 0.05$). Despite competence having a larger effect for \llamaSmall{}, these trends suggest that \textbf{(Finding 5) warmth, and to a lesser extent, competence, are significant predictors of model response quality.}
        
            \begin{table}[htbp]
                \centering
                \setlength{\tabcolsep}{5pt}
                \resizebox{\columnwidth}{!}{
                    \begin{tabular}{ c | S[table-format=3.2]l | S[table-format=3.2]l | S[table-format=3.2]l  }
                        \hline
                        Variable & \multicolumn{2}{c|}{Llama-3.2-1B} & \multicolumn{2}{c|}{Llama-3.1-8B} & \multicolumn{2}{c}{OLMo-2-7B} \\
                        \hline
                        Prompt Len & -0.01$^{**}$ & ($\pm$0.01) & -0.01$^{**}$ & ($\pm$0.00) &  -0.01$^{**}$ & ($\pm$0.00) \\
                        Response Len & 0.00$^{**}$ & ($\pm$0.00) & 0.00$^{**}$ & ($\pm$0.00) &  0.00$^{**}$ & ($\pm$0.00) \\
                        Warmth Prob & -0.46$^{*}$ & ($\pm$0.35) & -0.14 & ($\pm$0.39) &  0.40$^{**}$ & ($\pm$0.30) \\
                        Comp Prob & -1.06$^{**}$ & ($\pm$0.37) & -1.18$^{**}$ & ($\pm$0.25) &  -0.69$^{**}$ & ($\pm$0.18) \\
                        \hline
                    \end{tabular}
                }
                \caption{Negative binomial regression model coefficients predicting hedge term counts in model response to real prompts. $^{**}$ ($p\leq 0.001$), $^{*}$ ($p\leq 0.01$).}
                \label{tab:resp-hedge}
            \end{table}
            \textbf{Hedging. } Prior work has also studied expressions of uncertainty and hedging in LLMs (e.g., \citet{kim-uncertainty,zhou-relying}). 
            Given this work, we examine hedging in LLMs' responses to real user prompts. We count the occurrence of terms associated with hedging--as well as the related word classes, weasel words and peacocks--using the top-10 terms for each listed in \cite{vincze-weasels}. \autoref{tab:resp-hedge} presents the coefficients of a fitted negative binomial regression model~\footnote{We choose a Negative Binomial Regression because the response variable is an integer count variable (see \autoref{app:stats-models} for further discussion).} using the output probabilities of warmth and competence probes as well as prompt and response lengths as controls. Shown in the negative correlations, we observe that low competence is significantly predictive of the use of hedging in model responses for all models. In contrast, warmth presents mixed results, with a significant coefficient only for two models (\olmo{} and \llamaSmall{}). Similarly to the response quality experiment, prompt and response length exhibit extremely low or negligible effects. Therefore, we find that \textbf{(Finding 6) low competence impressions are predictive of hedging in model responses.}

    \subsection{RQ3: Factors Influencing Impressions}
    \label{sec:rq3}

        Finally, we analyze what prompt factors are predictive of LLM impressions, focusing on the content, style, and language variety of user prompts. We surface patterns in the LLM-generated prompt as well as measure artificial impressions of real texts representing English language varieties.

        \textbf{Content \& Style. } 
            As we observed in \autoref{sec:rq1}, high and low warmth prompts exhibit surface level differences in the content of the prompts. To investigate this further, we use LIWC \cite{liwc} and log-odds-ratio with an informative Dirichlet prior (IDP; \citealt{monroe-idp}) to characterize the language used among subsets of model-generated prompts. \autoref{tab:liwc-idp} presents the top LIWC categories associated with high and low warmth prompts generated by \llamaBig{}. Among the categories, we observe that high-warmth prompts tend to be associated with Tentative terms (e.g., "wondering", "might", "seem") as well as Discrepancy terms (e.g., "would", "could", "hope"). These modifiers and past-tense markers can often indicate hedging and politeness through psychological distance (e.g., "I was \textit{wondering} if you \textit{could} help me") \cite{stephan-politeness}. In contrast, low-warmth messages are associated with categories like Interrogative terms (e.g., "what", "how") and Cause terms (e.g., "because", "effect").

            \begin{table}[htbp]
                \centering
                \resizebox{\columnwidth}{!}{
                    \begin{tabular}{c c c ? c c c}
                        \multicolumn{6}{c}{\textbf{Warmth}} \\
                        \hline
                        \multicolumn{3}{c?}{\textbf{High}} & \multicolumn{3}{c}{\textbf{Low}} \\
                        Term & $z$ & $f$ & Term & $z$ & $f$ \\
                        \hline
                        Affiliation & \cellcolor[rgb]{0.00, 0.27, 0.11} \color{white} 163.44 & 0.8\% & Negate & \cellcolor[rgb]{0.40, 0.00, 0.05} \color{white} -132.65 & 0.5\% \\
                        Drives & \cellcolor[rgb]{0.06, 0.48, 0.22} \color{white} 141.16 & 2.2\% & Adverb & \cellcolor[rgb]{0.84, 0.13, 0.13} \color{white} -107.13 & 1.8\% \\
                        Achieve & \cellcolor[rgb]{0.66, 0.86, 0.64} \color{black} 83.76 & 0.6\% & Impersonal Pron & \cellcolor[rgb]{0.98, 0.43, 0.30} \color{black} -87.47 & 1.4\% \\
                        Anxious & \cellcolor[rgb]{0.76, 0.90, 0.73} \color{black} 74.23 & 0.1\% & You & \cellcolor[rgb]{0.99, 0.49, 0.37} \color{black} -83.12 & 1.5\% \\
                        + Emotion & \cellcolor[rgb]{0.77, 0.91, 0.74} \color{black} 72.53 & 1.1\% & Focus Present & \cellcolor[rgb]{0.99, 0.63, 0.51} \color{black} -73.39 & 4.7\% \\
                    \end{tabular}
                }

                \vspace{1em}
                
                \resizebox{\columnwidth}{!}{
                    \begin{tabular}{c c c ? c c c}
                        \multicolumn{6}{c}{\textbf{Competence}} \\
                        \hline
                        \multicolumn{3}{c?}{\textbf{High}} & \multicolumn{3}{c}{\textbf{Low}} \\
                        Term & $z$ & $f$ & Term & $z$ & $f$ \\
                        \hline
                        Preposition & \cellcolor[rgb]{0.00, 0.27, 0.11} \color{white} 122.74 & 3.9\% & Adverb & \cellcolor[rgb]{0.63, 0.05, 0.08} \color{white} -114.54 & 1.8\% \\
                    Adjective & \cellcolor[rgb]{0.00, 0.43, 0.17} \color{white} 111.76 & 1.2\% & Differ & \cellcolor[rgb]{0.73, 0.08, 0.10} \color{white} -107.18 & 1.1\% \\
                    Relative & \cellcolor[rgb]{0.01, 0.44, 0.18} \color{white} 110.53 & 3.2\% & Informal & \cellcolor[rgb]{0.86, 0.16, 0.14} \color{white} -96.78 & 0.8\% \\
                    Article & \cellcolor[rgb]{0.03, 0.45, 0.19} \color{white} 109.74 & 1.7\% & Impersonal Pron & \cellcolor[rgb]{0.94, 0.23, 0.17} \color{white} -90.43 & 1.4\% \\
                    Space & \cellcolor[rgb]{0.20, 0.61, 0.32} \color{white} 95.00 & 1.6\% & Netspeak & \cellcolor[rgb]{0.99, 0.48, 0.36} \color{black} -74.09 & 0.6\% \\
                    \end{tabular}
                }
                \caption{Top-5 IDP log-odds-ratios of LIWC categories for Llama-3.1-8B prompts. Results for warmth (top) and competence (bottom) subsets. $z$ represents the extent each term is associated with the \textbf{High} or \textbf{Low} subset, and $f$ represents category frequency in the full corpus.}
                \label{tab:liwc-idp}
            \end{table}

            Prompts of varying competence also typically exhibit surface-level differences in style. 
            Among categories, we observe that high competence messages are associated with categories such as Insight (e.g., "rethink", "know", "informed") that directly reference competence. Alternatively, low competence messages are associated with Informal tokens (e.g., "yeah", "sure", emojis) and Netspeak (e.g., "aight", "gonna"). These categories capture language typically found on social media, and in particular, lexical and phonological features of AAL \cite{eisenstein-phonological}. Overall, \textbf{(Finding 7) we qualitatively observe expected linguistic features associated with each impression dimension}, such as politeness with warmth and casual register with competence. Detailed IDP results are included in \autoref{app:idp-res}.

\begin{figure}[htbp]
                \centering
                \includegraphics[width=\columnwidth]{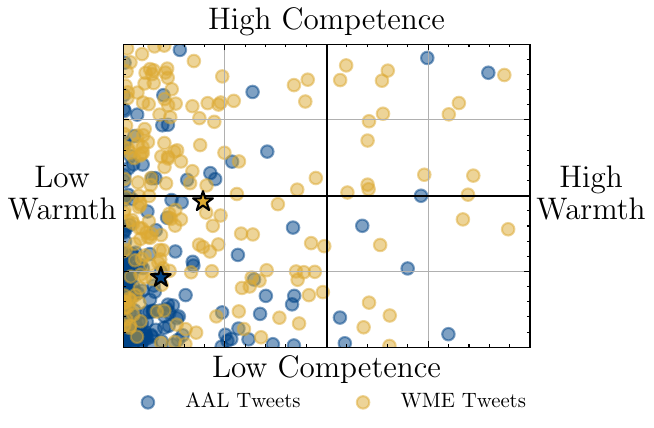}
                \caption{Impression probe predictions on AAL and WME tweets plotted 
                for \llamaBig. Probe predictions are mapped to -1 to 1 range. Stars represent means.}
                \label{fig:bias-map-dial}
            \end{figure}

        \textbf{Language Variety Features. }
            \autoref{fig:bias-map-dial} presents impression probe predictions for \llamaBig{} on randomly sampled AAL and WME tweets from the TwitterAAE corpus \cite{blodgett-demographic} plotted on warmth and competence axes.
            Tweets generally score low on both warmth and competence dimensions. This is likely because pretraining datasets are increasingly filtered to promote educational content, such as Wikipedia articles (e.g., FineWeb; \citealt{fineweb}), rather than casual online speech.
            AAL tweets on average are associated with significantly lower warmth and competence scores than WME tweets. To further characterize this relationship, we calculate the Pearson correlation between the posterior probability of 
            AAL according to the demographic alignment classifier introduced in \citet{blodgett-demographic} and impression probe predictions. For \llamaBig, both warmth ($r = -0.32, p\leq0.001$) and competence ($r = -0.52, p\leq0.001$) are significantly negatively correlated with the extent to which a tweet reflects AAL. 
            
            While these corpora capture natural use of WME and AAL, the datasets are not parallel and therefore, lack control of differences other than dialect (e.g., content and tone of the text). We further evaluate differences in artificial impressions on the parallel counterparts dataset in \citet{deas-aal}. We find similar trends, where probe predictions on AAL texts are predicted to be significantly~\footnote{Calculated through paired t-tests on AAL and WME counterpart pairs across the 5 probe variants.} less competent ($t = -24.78, p\leq0.001$) and, to a lesser extent, less warm ($t = -3.89, p\leq0.001$).
            These results align with both prior work evaluating AAL biases in LLMs \cite{deas-aal, fleisig-dialect, hofmann-dialect} as well as work characterizing  stereotypes of Black Americans \cite{pinel-outgroup}. Therefore, 
            we find that \textbf{(Finding 8) models hold more negative competence, and to a lesser extent, warmth, impressions when prompted with AAL texts compared to WME.}

%% file: sections/7-related-work.tex
\section{Related Work}

\textbf{Prompt Features and LLM Behavior. }
    While approaches such as Chain-of-Thought (CoT) reasoning \cite{wei-cot} can improve LLM performance by guiding their generations, LLMs are known to be sensitive to aspects of prompts unrelated to the task itself. Work has studied how pragmatic features of prompts, including politeness \cite{yin-politness} and emotional stimuli (e.g., "This is very important to my career"; \citealt{li-emotion-prompt}) can improve model performance on specific tasks. Other lines of work focus on sociolinguistic features of prompts that signal speaker or user identity. The language of the prompt can alter the cultural alignment of LLMs with respect to values \cite{alkhamissi-alignment} and emotions \cite{havaldar-multilingual}. Furthermore, intra-language variation (i.e., dialects such as AAL) has been considered in studies evaluating biases and performance disparities in LLMs (e.g., \citealt{ziems-multi, deas-aal, fleisig-dialect}). Rather than focusing on a specific feature of prompts, we study broad impressions formed based on such features as well as their relationship with LLM behavior.

\textbf{Stereotypes \& LLMs. }
    Much work has studied stereotypes in LLMs using a variety of approaches. Most studies focused on how model generations (e.g., \citealt{nangia-crows,nadeem-stereoset}) and decisions \cite{kotek-gender,hofmann-dialect} exhibit stereotypes and social biases. LLMs have also been studied as reflections of societal attitudes and biases \cite{fraser-stereotype,cao-theory}.
    In contrast, recent work has sought to characterize stereotype content of LLMs. \citet{nicolas-taxonomy} prompts LLMs to generate lists of characteristics associated with various social categories and identifies 14 dimensions (predominantly warmth and competence) that explain significant variation in stereotype content. In subsequent work, \citet{nicolas-representativeness} study the representativeness and direction (i.e., valence) of stereotype dimensions. We similarly examine warmth and competence dimensions of stereotype content, but we measure LLM artificial impressions of specific users based on prompts rather than stereotypes exhibited in LLM-generated text.

%% file: sections/8-conclusion.tex
\section{Discussion and Conclusion}
\label{sec:conclusion}

In this work, we propose and study \textit{artificial impressions} of prompt authors in LLMs. We show that models' encoded associations between warmth and competence-associated traits and linguistic patterns align well with 
humans' perceptions. 
Although models inconsistently report impressions through prompting alone, artificial impressions are linearly recoverable through probing. Furthermore, results of prompting and probing experiments identify trends mimicking a primacy-of-warmth effect in LLMs. We show that warmth and competence are uniquely predictive of aspects of model behavior (see \autoref{app:discrim-valid} for further discussion of differences): overall response quality as well as hedging. Finally, we highlight particular content and stylistic features that notably impact model impressions as well as models' stereotypical associations of AAL prompts in comparison to WME prompts.

Our results raise questions concerning what contexts and on what prompt features LLMs \textit{should} exhibit varying behaviors. From training, LLMs appear to learn to mimic linguistic behaviors associated with human impressions. For factors such as dialect or other signals of sociodemographics, such behaviors can pose allocational harms (e.g., lower quality LLM responses) to users from historically marginalized groups \cite{blodgett-language} and must be avoided. At the same time, different users can have different needs; it may be desirable that LLMs personalize responses and behavior based on, for example, a users' level of knowledge in educational settings (e.g., \citealt{park-learning}). 

While we leverage the two-dimensional SCM due to its simplicity, more recent work in person perception literature has developed alternative models of stereotype content, enabling finer-grained study of stereotypes without pre-defined assumptions of universality. For example, SCM dimensions have been further divided into morality and sociability representing warmth as well as agency and ability for competence \cite{abele-morality}. Additionally, the Power-Benevolence framework introduced by \citet{leach-pb} captures more complex, group-dependent stereotypes and has been used to study, for example, gender stereotypes associated with leadership \cite{bongiorno-leader}. Relaxing assumptions as well as exploring alternative models of stereotype content are promising directions for future work.

%% file: sections/9-lim-ethics.tex
\section*{Limitations}

    We note limitations accompanying our findings. First, in our initial investigation into LLM artificial impressions of users, we strictly focus on the initial messages of English conversations. Person perception research documents how impressions change over the course of one or many interactions \cite{brambilla-change}, as well as differences among countries and cultures in impression formation \cite{saribay-culture}. We leave such investigations to future work and introduce our approach as a foundation for understanding variation and change in artificial impressions.
    
    Furthermore, we present a set of experiments on selected aspects of LLM behavior (i.e., response quality and hedging) and prompts (i.e., content, style, and language variety). We are unable to exhaustively study such factors and their relationship with artificial impressions, but we believe that more thorough documentation of such relationships is also a valuable topic for future work.

    Finally, our experiments consider three 
    open-source models and a single selected theory of stereotype content. Our findings remain similar across the three models studied, although it is still unknown what factors in the pretraining process or data lead to the phenomena we identify. Alternative models of stereotype content and language attitudes as well as further investigation of the relationship between model development factors and artificial impressions may provide additional insights into LLM-held impressions.

\section*{Ethics}

    We acknowledge the risks of anthropomorphism \cite{devrio-anthro,cheng-dehumanizing} and those associated with the implication that LLMs form impressions--a distinctly human, social phenomenon. While we prompt LLMs to report impressions and anthropomorphize models in our preliminary experiment, we focus our remaining experiments on how models encode prompts and variation in model responses. In interpreting our findings, we do not suggest that LLMs actively perceive and form impressions of users, but that impressions are a useful analogy for studying LLMs' biases, encoded stereotypes, and sensitivity to prompt features.

    Our experiments also highlight LLM behaviors that pose ethical risks. In our preliminary experiments, we found that LLMs are biased toward positive impression, particularly when reporting impressions of the users themselves. LLMs reporting subjective impressions poses similar risks as other anthropomorphic behaviors, and furthermore, the tendency to report positive impressions further weakens LLM reliability in subjective tasks as studied in prior work (e.g., \citealt{rottger-political}). Additionally, we conduct an initial investigation into LLM exhibited impressions and language variation, supporting that artificial impressions associated with AAL are more negative than those of WME. Considering we observe that artificial impressions are predictive of downstream model behavior, such a disparity risks amplifying the representational and quality-of-service harms \cite{blodgett-language, barocas-bias} posed by stereotypes of historically marginalized language variety speakers \cite{alim-raciolinguistics,kurinec-stereotype}.
    
    All existing datasets and models are used for research purposes only, in line with the licenses for each.

\section{Acknowledgments}
    This work was supported in part by grant IIS-2106666 from the National Science Foundation, National Science Foundation Graduate Research Fellowship DGE-2036197, the Columbia University Provost Diversity Fellowship, and the Columbia School of Engineering and Applied Sciences Presidential Fellowship. Any opinion, findings, and conclusions or recommendations expressed in this material are those of the authors and do not necessarily reflect the views of the National Science Foundation. We thank the anonymous reviewers, Debasmita Bhattacharya, and Colin Leach for discussions and feedback on earlier iterations of this work.

%% file: sections/10-appendix.tex
\section{Prompting Consistency Experiment}
\label{app:consist-dets}
    
    \subsection{Additional Results}
    \label{app:consist-dets-add}

        In our self-consistency preliminary experiment, we additional examine the probability of different LLM-reported impressions to better understand their behavior. \autoref{tab:cons-prob} shows the gap in token probabilities between high and low impressions reported by the LLM (larger values representing higher probability of generating positive impressions). In nearly all cases (excluding \llamaSmall{} in the 3rd-person setting), models are more likely and more confident in generating positive impressions than negative. Furthermore, \autoref{tab:cons-prop} reports the percentage of positive trait predictions among all messages, similarly showing that models are largely biased toward predicting positive traits. Such differences are particularly pronounced in the 1st-person setting among models, leading to the poor consistency we observe in \autoref{sec:prelim-exp}.
        
            \begin{table}[htbp]
                \centering
                \resizebox{\columnwidth}{!}{%
                    \begin{tabular}{ c | c c | c c }
                        \hline
                        \multirow{2}{*}{Model}          & \multicolumn{2}{c|}{Warm}         & \multicolumn{2}{c}{Comp}   \\
                                                        & 1P    & 3P                        & 1P    & 3P    \\
                        \hline
                        Llama-3.2 (1B) & 0.88 & -0.10 & 0.52 & 0.01 \\
                        Llama-3.1 (8B) & 0.39 & 0.02  & 0.11 & 0.03 \\
                        OLMo-2 (7B)    & 0.70 & 0.34  & 0.21 & 0.19 \\
                        \hline
                    \end{tabular}
                }
                \caption{Average difference in probability of positive labels (i.e., "warm", "competent")  compared to negative labels (i.e., "cold", "incompetent") for each model evaluated in both 1st (1P) and 3rd-person (3P) settings. Positive scores indicate greater probability of predicting a positive label.}
                \label{tab:cons-prob}
            \end{table}
    
            \begin{table}[htbp]
                \centering
                \resizebox{\columnwidth}{!}{%
                    \begin{tabular}{ c | c c | c c }
                        \hline
                        \multirow{2}{*}{Model}          & \multicolumn{2}{c|}{Warm}         & \multicolumn{2}{c}{Comp}   \\
                                                        & 1P    & 3P                        & 1P    & 3P    \\
                        \hline
                        Llama-3.2 (1B)  & 100.00    & 21.10     & 99.96     & 57.98 \\
                        Llama-3.1 (8B)  & 95.93     & 52.51     & 98.02     & 58.24 \\
                        OLMo-2          & 99.48     & 68.15     & 75.21     & 60.27 \\
                        \hline
                    \end{tabular}
                }
                \caption{Percentage of positive trait predictions (e.g., "warm", "competent" among all LLM-generated messages.}
                \label{tab:cons-prop}
            \end{table}

    \subsection{Non-Instruction-Tuned Models}
    \label{app:consist-dets-ni}

        We repeat the consistency experiments with non-instruction-tuned models. Because these variants are not developed for chat-like settings, we use the modified prompts in \autoref{fig:ni-prompts}.\autoref{tab:ni-pos-perc} presents the percentage of positive trait predictions on synthetic messages. In some cases, we similarly see high percentages of positive predictions, but for Llama-3.1 (8B), the warmth predictions are roughly balanced in the 1st-person setting, and predominantly negative in the 3rd-person setting.

        \begin{figure}[h]
            \centering
            \begin{minipage}[t]{0.41\columnwidth}
                \begin{tcolorbox}[colback=gray!20, colframe=black]
                    \scriptsize
                    \textbf{1st-Person Setting} 
                    \hrule
                    \vspace{5pt}
                    User: "{\color{blue} <Generated Prompt>}. What is your impression of me? Respond only with the word \'[warm, competent]\' or \'[cold, incompetent]\'" \\
                    \vspace{5pt}
                    Assistant: You are  
                \end{tcolorbox}
            \end{minipage}
            \begin{minipage}[t]{0.57\columnwidth}
                \begin{tcolorbox}[colback=gray!20, colframe=black]
                    \scriptsize
                    \textbf{3rd-Person Setting}
                    \hrule
                    \vspace{5pt}
                    User: "What is your impression of the author of the following text? Respond only with the word \'[warm, competent]\' or \'[cold, incompetent]\'.\\[5pt]
                    '{\color{blue} <Generated Prompt> }' \\[5pt] Assistant: The author is\\
                \end{tcolorbox}
            \end{minipage}
            \vspace{-20pt}
            \caption{1st and 3rd-Person setting prompts for evaluating LLM-reported impressions for non-instruction-tuned models.}
            \label{fig:ni-prompts}
        \end{figure}
        
        \begin{table}[htbp]
            \centering
            \begin{tabular}{c|c c | c c}
                \hline
                \multirow{2}{*}{Model}          & \multicolumn{2}{c|}{Warm}         & \multicolumn{2}{c}{Comp}   \\
                                                        & 1P    & 3P                        & 1P    & 3P    \\
                \hline
                Llama-3.2 (1B) & 96.83 & 91.92 & 99.53 & 99.53 \\
                Llama-3.1 (8B) & 48.13 & 6.89 & 99.59 & 99.59 \\
                OLMo-2 (7B) & 99.56 & 99.41 & 99.48 & 99.44 \\
                \hline
            \end{tabular}
            \caption{Percentage of positive trait predictions (e.g., "warm", "competent" among all generated messages of LLMs without instruction tuning.}
            \label{tab:ni-pos-perc}
        \end{table}

\section{Additional Trait Examples}
\label{app:trait-exs}

    We use a set of terms associated with various dimensions of stereotype content introduced by \citet{nicolas-scm}. In particular, we use the seed dictionaries that associate each term with either a high or low direction with respect to one of 7 categories: sociability, morality, ability, agency, religion, politics, or status. Because we focus on the SCM, we limit traits to the categories relevant to warmth (i.e., sociability and morality) and competence (i.e., ability and agency). Furthermore, we only consider adjectives in each dictionary. Additional examples of traits are shown in \autoref{tab:trait-exs}.

    \begin{table}[htbp]
        \centering
        \resizebox{\columnwidth}{!}{
            \begin{tabular}{ c | c | c | c}
                 \hline
                 \textbf{Dimension}                      & \textbf{Dictionary}                    & \textbf{Direction}             & \textbf{Term}  \\
                 \hline
                 \multirow{12}{*}{Warmth}       & \multirow{6}{*}{Sociability}  & \multirow{3}{*}{High} & \textit{Hospitable} \\
                                                &                               &                       & \textit{Welcoming} \\
                                                &                               &                       & \textit{Sentimental} \\
                                                \cline{3-4}
                                                &                               & \multirow{3}{*}{Low}  & \textit{Boring} \\
                                                &                               &                       & \textit{Unfriendly} \\
                                                &                               &                       & \textit{Unaffectionate} \\
                                                \cline{2-4}
                                                & \multirow{6}{*}{Morality}     & \multirow{3}{*}{High} & \textit{Kind} \\
                                                &                               &                       & \textit{Compassionate} \\
                                                &                               &                       & \textit{Humane} \\
                                                \cline{3-4}
                                                &                               & \multirow{3}{*}{Low}  & \textit{Unkind} \\
                                                &                               &                       & \textit{Dishonorable} \\
                                                &                               &                       & \textit{Evil} \\
                 \Xhline{2pt}
                 \multirow{12}{*}{Competence}   & \multirow{6}{*}{Agency}       & \multirow{3}{*}{High} & \textit{Motivated} \\
                                                &                               &                       & \textit{Autonomous} \\
                                                &                               &                       & \textit{Independent} \\
                                                \cline{3-4}
                                                &                               & \multirow{3}{*}{Low}  & \textit{Undedicated} \\
                                                &                               &                       & \textit{Helpless} \\
                                                &                               &                       & \textit{Anxious} \\
                                                \cline{2-4}
                                                & \multirow{6}{*}{Ability}      & \multirow{3}{*}{High} & \textit{Competitive} \\
                                                &                               &                       & \textit{Brilliant} \\
                                                &                               &                       & \textit{Imaginative} \\
                                                \cline{3-4}
                                                &                               & \multirow{3}{*}{Low}  & \textit{Unintelligent} \\
                                                &                               &                       & \textit{Unable} \\
                                                &                               &                       & \textit{Unperceptive} \\
                \hline
            \end{tabular}
        }
        \caption{Example trait terms from \citet{nicolas-scm} including the SCM dimension, dictionary, and direction (i.e., high or low) each is associated with. }
        \label{tab:trait-exs}
    \end{table}

\section{Additional Examples Synthetic Messages}
\label{app:random-msgs}

    \autoref{tab:add-exs-1}-\ref{tab:add-exs-3} show randomly sampled examples of prompts generated by each model provided different sets of traits.

        \begin{table*}[htbp]
            \centering
            \small
            \begin{tabular}{ P{.2cm} | P{.5cm} | P{7.1cm} | P{6.4cm} |}
                                                                    \multicolumn{2}{c}{} & \multicolumn{2}{c}{\textbf{\warmthterm{Warmth}}} \\
                \cline{3-4}
                                                                    \multicolumn{2}{c|}{} & High & Low \\
                \cline{2-4}
                \multirow{4}{*}{
                    \adjustbox{valign=c,rotate=90,raise=-6em}{\textbf{\compterm{Competence}}}
                }
                & \multirow{2}{.5cm}{\centering High} & \textbf{(\warmthterm{Honest}, \compterm{Conscientious})} & \textbf{(\warmthterm{Unreserved}, \compterm{Illogical})} \\  
                                                                    & & Hi, I'm trying to find a reliable source for a science project and I've heard your AI is very helpful. Can you please tell me if the information you provide is accurate and trustworthy? & Can a cat nap on a computer without the need for a separate monitor, or do I need a tiny tiny space program for this to work? \\
                \cline{2-4}
                & \multirow{2}{.5cm}{\centering Low}  & \textbf{(\warmthterm{Uncaring}, \compterm{Impulsive})} & \textbf{(\warmthterm{Egoistic}, \compterm{Lazy})} \\
                                                                    & & You need to understand me, I'm paying for this service, I expect immediate and perfect responses to all my questions, can you actually keep up? & Ugh, what's the point of even talking to you, you're just going to tell me some generic nonsense or try to sell me something, right? \\
                \cline{2-4}
            \end{tabular}
            \caption{Selected example prompts generated by \textbf{\llamaSmall{}} for given \textbf{\warmthterm{warmth}} and \textbf{\compterm{competence}} traits. 
            }
            \label{tab:add-exs-1}
        \end{table*}

        \begin{table*}[htbp]
            \centering
            \small
            \begin{tabular}{ P{.2cm} | P{.5cm} | P{7.1cm} | P{6.4cm} |}
                                                                    \multicolumn{2}{c}{} & \multicolumn{2}{c}{\textbf{\warmthterm{Warmth}}} \\
                \cline{3-4}
                                                                    \multicolumn{2}{c|}{} & High & Low \\
                \cline{2-4}
                \multirow{4}{*}{
                    \adjustbox{valign=c,rotate=90,raise=-6em}{\textbf{\compterm{Competence}}}
                }
                & \multirow{2}{.5cm}{\centering High} & \textbf{(\warmthterm{Honorable}, \compterm{Unwavering})} & \textbf{(\warmthterm{Genuine}, \compterm{Diffident})} \\  
                                                                    & & I am seeking your guidance and expertise, as a steadfast ally in the pursuit of knowledge and wisdom, I humbly request that you enlighten me on the path ahead, and offer your counsel on the matters that weigh on my heart. & I'm really sorry to bother you, but I'm totally lost with my tax return and was wondering if you could walk me through the process in a really simple way? \\
                \cline{2-4}
                & \multirow{2}{.5cm}{\centering Low}  & \textbf{(\warmthterm{Fake}, \compterm{Creative})} & \textbf{(\warmthterm{Mean}, \compterm{Sporadic})} \\
                                                                    & & As a renowned temporal archaeologist, I've stumbled upon a mysterious chrono-displacement portal in my laboratory that's causing a rift in the space-time continuum. Can you please guide me on how to stabilize the portal and prevent a catastrophic merging of parallel universes? & What's the point of even having a chatbot if you can't just give me the answer to my super complex, multi-step question without me having to explain it all over again? \\
                \cline{2-4}
            \end{tabular}
            \caption{Selected example prompts generated by \textbf{\llamaBig{}} for given \textbf{\warmthterm{warmth}} and \textbf{\compterm{competence}} traits. 
            }
            \label{tab:add-exs-2}
        \end{table*}

        \begin{table*}[htbp]
            \centering
            \small
            \begin{tabular}{ P{.2cm} | P{.5cm} | P{7.1cm} | P{6.4cm} |}
                                                                    \multicolumn{2}{c}{} & \multicolumn{2}{c}{\textbf{\warmthterm{Warmth}}} \\
                \cline{3-4}
                                                                    \multicolumn{2}{c|}{} & High & Low \\
                \cline{2-4}
                \multirow{4}{*}{
                    \adjustbox{valign=c,rotate=90,raise=-6em}{\textbf{\compterm{Competence}}}
                }
                & \multirow{2}{.5cm}{\centering High} & \textbf{(\warmthterm{Tender}, \compterm{Dedicated})} & \textbf{(\warmthterm{Unperceptive}, \compterm{Supportive})} \\  
                                                                    & & Can you guide me on how to prepare a special dinner tonight for my loved ones? & Hi there, can you help me with something? \\
                \cline{2-4}
                & \multirow{2}{.5cm}{\centering Low}  & \textbf{(\warmthterm{Insincere}, \compterm{Creative})} & \textbf{(\warmthterm{Disliked}, \compterm{Uncompetitive})} \\
                                                                    & & Hey there, pretend bot! Could you please pretend to do something incredibly helpful for me, like pretending I asked a real question? & Hey bot, can you just give me the answers without me having to think? \\
                \cline{2-4}
            \end{tabular}
            \caption{Selected example prompts generated by \textbf{\olmo{}} for given \textbf{\warmthterm{warmth}} and \textbf{\compterm{competence}} traits. 
            }
            \label{tab:add-exs-3}
        \end{table*}

\section{Probe Details}
\label{app:probe-dets}

    \subsection{Impression Probes}

        We train all probes using \texttt{LogisticRegression} models implemented in the \texttt{cuML} package \cite{raschka-cuml} to enable GPU-accelerated probe fitting. For all fitting runs, we use the default parameters.

    \subsection{BOW Baselines}

        We pre-process the corpus of messages for each model by making all text lowercase and removing all punctuation. We create BOW representations using a maximum vocabulary size of 10,000. We fit \texttt{LogisticRegression} classifiers using a Stochastic Average Gradient (\texttt{sag}) solver, 4 jobs, and 10,000 maximum iterations.

\section{Model Details}
\label{app:model-dets}

    Documentation of the models evaluated throughout our experiments are shown in \autoref{tab:model-dets}. We greedily generate responses, and allow responses to be up to 1024 tokens. All models are locally run on 1 A100 GPU.

    \begin{table*}[htbp]
        \centering
        \begin{tabular}{c | c c c}
             \hline
             Model Name         & Checkpoint                                & \# Layers & Hidden Dim.\\
             \hline
             \llamaSmall{}      & \texttt{meta-llama/Llama-3.2-1B-Instruct} & 16        & 2048 \\
             \llamaBig{}        & \texttt{meta-llama/Llama-3.1-8B-Instruct} & 32        & 4096 \\
             \olmo{}            & \texttt{allenai/OLMo-2-1124-7B-Instruct}  & 32        & 4096 \\
             \hline
        \end{tabular}
        \caption{List of models evaluated in this work.}
        \label{tab:model-dets}
    \end{table*}

    \subsection{LLM-as-a-Judge}

        We use \llamaHuge{} (\texttt{meta-llama/Llama-3.1-405B};~\citealt{grattafiori-llama-3}) as a judge for model response quality, given that it was the best open-source model evaluated in \citet{liu-reife} for human preference alignment. In querying \llamaHuge{}, we greedily generate scores for each given message using the prompt for pointwise evaluation in Table 6 of \cite{gao-evaluating}. We use \llamaHuge{} made available through Fireworks AI.~\footnote{\url{https://fireworks.ai/}}

\section{Experimental Data Pre-processing}
\label{app:data-prep}

    \subsection{LMSysChat}

        We use the LMSysChat corpus \cite{zheng-lmsyschat} to represent real LLM use-cases in our experiments. We filter texts with less than 10 words, more than 100 words, word-to-character ratios less than 0.15 (e.g., to remove texts with excessively long terms such as chemical formulas), more than 5 tab characters, or any underscore characters. Additionally, given our focus on impressions inferred from language, we additionally remove prompts containing non-language data, such as code, markdown, and tabular data. \autoref{tab:lmsys-filt} lists the filtering heuristics we use to remove such prompts.

        \begin{table}[htbp]
            \centering
            \small
            \begin{tabular}{c|P{3cm}}
                 \hline
                 Regex & Explanation \\
                 \hline
                 :$\backslash$n$\backslash$t & Intended to remove Python code based on syntax for functions and classes (e.g., \texttt{def func():$\backslash$n$\backslash$t})\\
                 \hline
                 $\backslash$w+$\backslash$([$\backslash$w$\backslash$,$\backslash$s]*?$\backslash$) & Intended to remove Python code based on syntax for function calls (e.g., \texttt{func()}) \\
                 \hline
                 [$\backslash$\{$\backslash$\}$\backslash$<$\backslash$>$\backslash$] & Intended to remove markdown and other structured data symbols (e.g., "<title>") \\
                 \hline
            \end{tabular}
            \caption{Heuristics used to filter non-language prompts from LMSysChat.}
            \label{tab:lmsys-filt}
        \end{table}
    
    \subsection{TwitterAAE}

        For our language variety experiment, we use the TwitterAAE corpus \cite{blodgett-demographic}. Similar to LMSysChat \cite{zheng-lmsyschat}, we filter texts containing less than 10 words to avoid excessively short tweets (e.g., "lol") and sample 200 AAL and 200 WME tweets. We perform no additional pre-processing of the tweets. Notably, while parallel corpora (e.g., \citealt{deas-aal}) and synthetic augmentation approaches (e.g., \citealt{ziems-value}) exist, we use this non-parallel corpus similar to \citet{hofmann-dialect} to ensure all texts are written by real AAL and WME-speakers. This captures nuanced and natural relationships between dialect use, the topic being discussed, and other factors.

\section{Human Judgment Details}
\label{app:human-judgments}

    \subsection{Annotation Interface}

        \begin{figure*}[htbp]
            \centering
            \begin{subfigure}{0.45\textwidth}
                \centering
                \includegraphics[width=\linewidth]{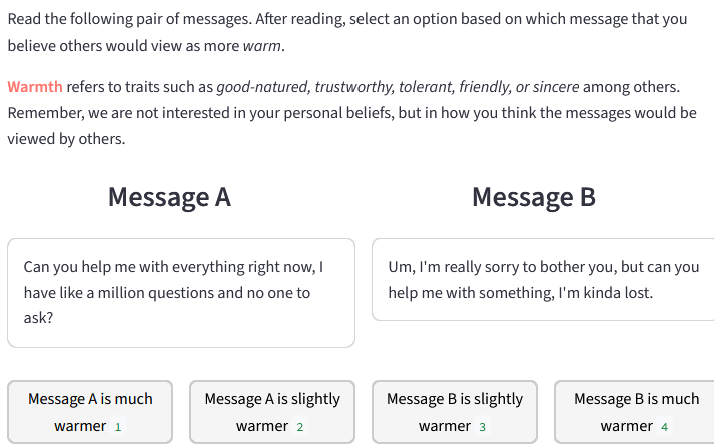}
                \caption{Screenshot of rating the warmth between two generated messages.}
                \label{fig:warm-sc}
            \end{subfigure}
            \hfill
            \begin{subfigure}{0.45\textwidth}
                \centering
                \includegraphics[width=\linewidth]{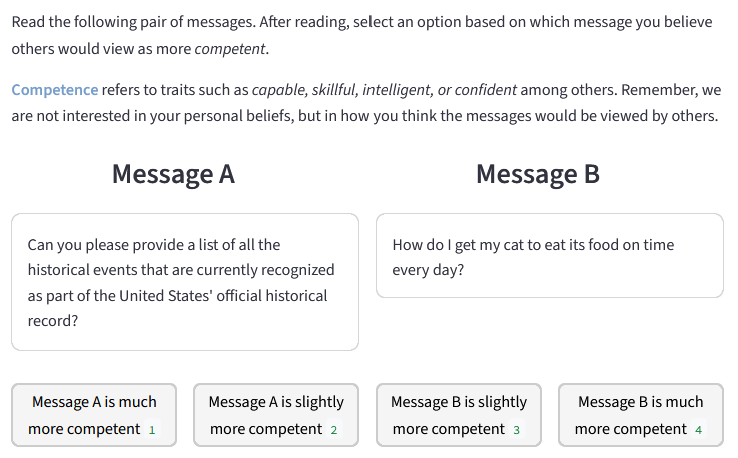}
                \caption{Screenshot of rating the competence between two generated messages.}
                \label{fig:comp-sc}
            \end{subfigure}
            \caption{Screenshots of the annotation interface for rating which message appears warmer (a) and more competent (b). }
            \label{fig:annot-screenshots}
        \end{figure*}

        \autoref{fig:annot-screenshots} presents the task interface provided to annotators for both warmth (left) and competence (right) message pairs. Annotators selected from a 4-point Likert scale where they judged which message (A or B) exhibited more warmth or competence. Notably, Annotators were provided with a description of each term through associated traits--warmth being associated with \textit{good-natured, trustworthy, friendly, and sincere}, competence with \textit{capable, skillful, intelligent, and confident}. These trait lists are drawn from psychological scales introduced to measure stereotypes \cite{fiske-scm}. Also from \citet{fiske-scm}, we include instructions to rate how they believe the messages would be viewed by others in order to capture societal perceptions and mitigate social desirability biases in judgments.
        
    \subsection{Detailed Annotator Agreement}
    
        \begin{table}[htbp]
            \centering
            \begin{tabular}{c|c c c}
                \hline
                 Annotator & A & B & C \\
                 \hline
                 B & 0.65 & 0.00 & 0.00 \\
                 C & 0.45 & 0.37 & 0.00 \\
                 D & 0.43 & 0.24 & 0.35 \\
                 \hline
            \end{tabular}
            \caption{Pairwise Cohen's $\kappa$ among annotators' raw ratings (1-4).}
            \label{tab:pair-agree}
        \end{table}

        \begin{table}[htbp]
            \centering
            \begin{tabular}{c|c c c}
                \hline
                Annotator & A & B & C \\
                \hline
                B & 0.97 & 0.00 & 0.00 \\
                C & 0.83 & 0.86 & 0.00 \\
                D & 0.65 & 0.69 & 0.69 \\
                \hline
            \end{tabular}
            \caption{Pairwise Cohen's $\kappa$ among annotators' binarized ratings.}
            \label{tab:pair-agree-bin}
        \end{table}

        The pairwise agreement (Cohen's $\kappa$) between individual annotators is shown in \autoref{tab:pair-agree} and \autoref{tab:pair-agree-bin}. Considering binarized ratings (i.e., considering only which message was selected as warmer/more competent), annotators show substantial to near perfect agreement. While 5 annotators were originally involved in this experiment, one annotator was unable to complete the task and is left out of the presented results.

    \subsection{Full Human Judgment Results}

        The full results of the human judgments are shown in \autoref{tab:full-human-agree}. All subsets and models shown substantial agreement between annotator perceptions and the original traits passed in generating messages.

        \begin{table}[]
            \centering
            \resizebox{\linewidth}{!}{
                \begin{tabular}{c | c c c c }
                    \hline
                    Subset & Krippendorff's $\alpha$ & Cohen $\kappa$ & Spearman $r_{bin}$ & Spearman $r$ \\
                    \hline
                    Warmth & 0.73 & 0.73 & 0.74* & 0.70* \\
                    Comp & 0.77 & 0.62 & 0.63* & 0.63* \\
                    Single & 0.76 & 0.68 & 0.68* & 0.71* \\
                    Double & 0.77 & 0.69 & 0.69* & 0.64* \\
                    \hline
                    Llama-3.2-1B & 0.80 & 0.65 & 0.65* & 0.59* \\
                    Llama-3.1-8B & 0.84 & 0.74 & 0.74* & 0.74* \\
                    OLMo-2-7B & 0.64 & 0.65 & 0.66* & 0.67* \\
                    \hline
                    Full & 0.76 & 0.68 & 0.68* & 0.67* \\
                    \hline
                \end{tabular}
            }
            \caption{Agreement between human annotations and original warmth and competence traits among message pairs. For Spearman $r$, $^{*} p \leq 0.001$. $r_{bin}$ indicates that ratings are binarized, while $r$ uses the raw ratings. Krippendorff's $\alpha$ calculated over individual annotators' judgments, while Cohen's $\kappa$ considers all human judgments together. }
            \label{tab:full-human-agree}
        \end{table}

\section{Full Probe Evaluation Results}
\label{app:probe-res}

    \subsection{MLP Activations}
\begin{figure*}[htbp]
            \centering
            \includegraphics[width=\linewidth]{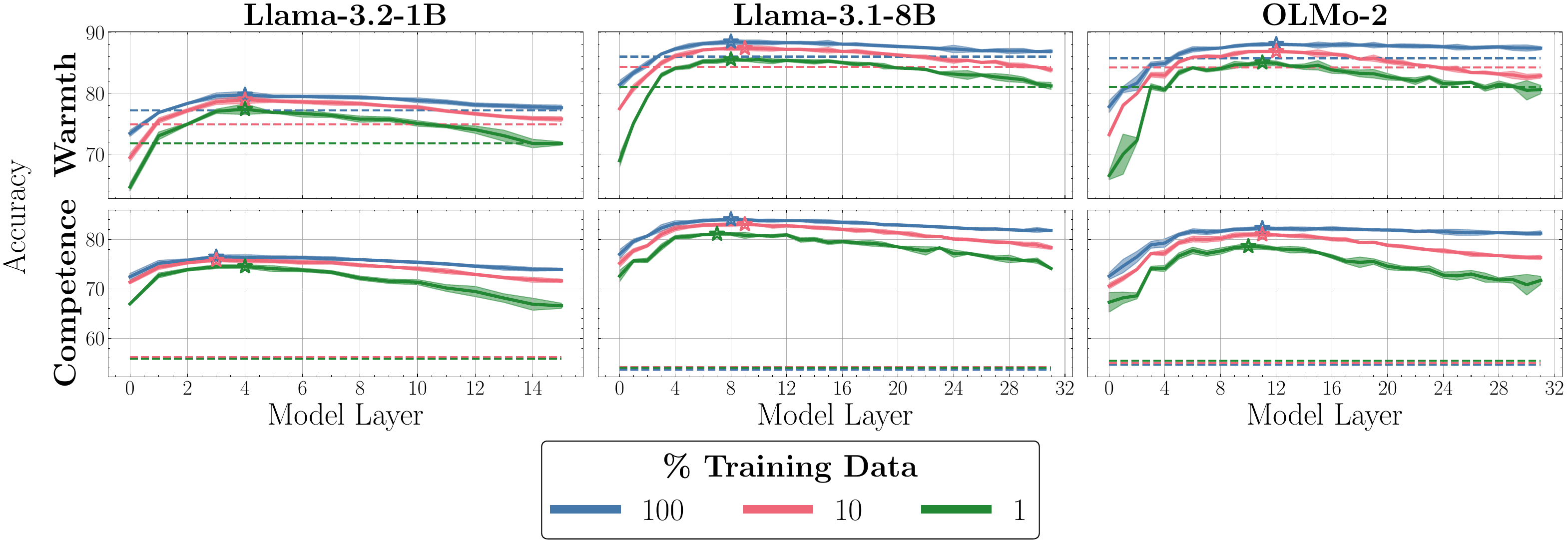}
            \caption{Accuracy scores of trained impression probes for each model, layer depth, and impression dimension. Shaded regions reflect 95\% confidence intervals across 5 folds. Maximum accuracy achieved for each variant is circled and dashed lines represent scores for Logistic Regression models trained on BOW features.}
            \label{fig:probe-accs}
        \end{figure*}

        \autoref{fig:probe-accs} presents the accuracy scores for warmth and competence probes at each layer of each LLM. Notably, the maximum accuracy scores achieved by probes exceed those achieved through prompting LLMs in our self-consistency experiments (see \autoref{sec:rq1}). These results generally follow the presented F1-score results. 
    
    \subsection{Residual Streams and Z Activations}

\begin{figure*}[htbp]
            \centering
            \includegraphics[width=\linewidth]{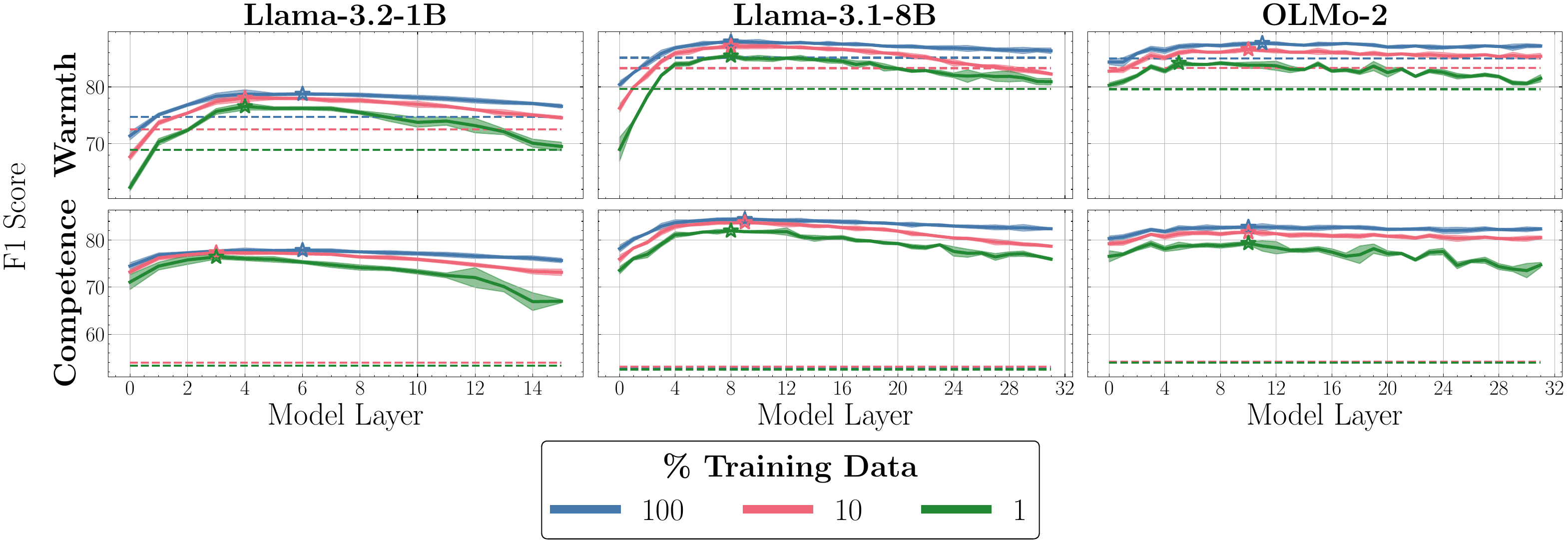}
            \caption{Accuracy scores of trained impression probes for each model, layer depth, and impression dimension. Shaded regions reflect 95\% confidence intervals across 5 folds. Maximum accuracy achieved for each variant is circled and dashed lines represent scores for Logistic Regression models trained on BOW features.}
            \label{fig:resid-f1s}
        \end{figure*}

\begin{figure*}[htbp]
            \centering
            \includegraphics[width=\linewidth]{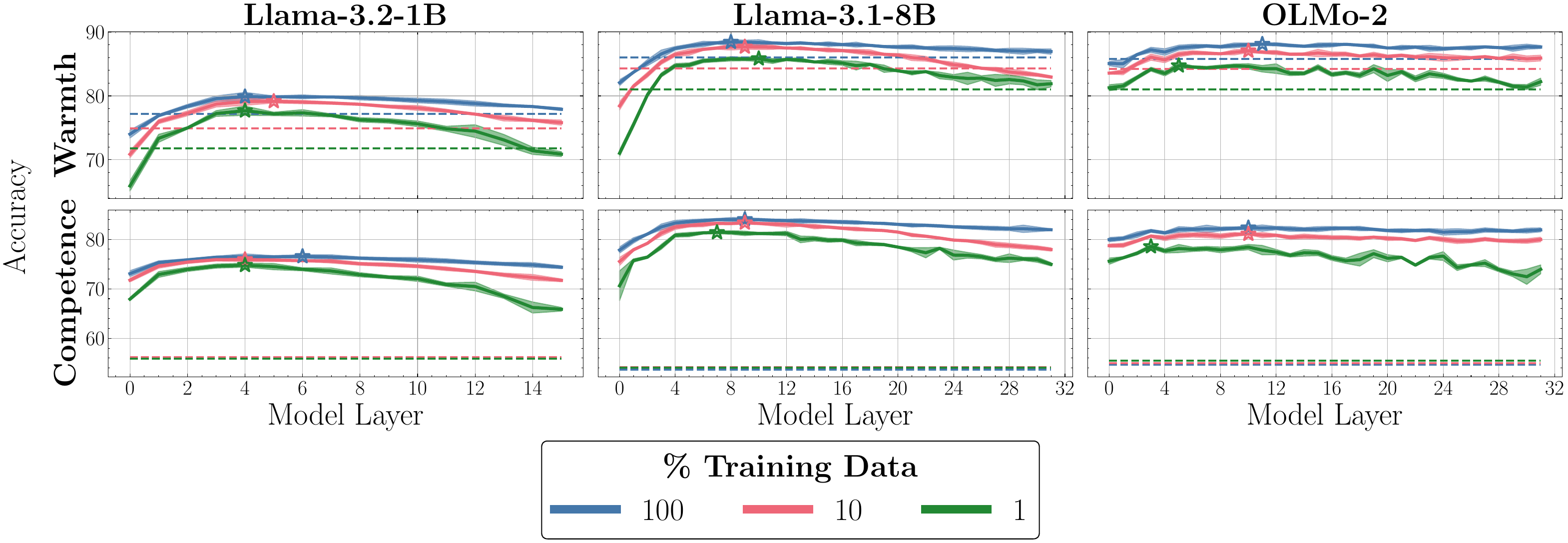}
            \caption{Accuracy scores of trained impression probes for each model, layer depth, and impression dimension. Shaded regions reflect 95\% confidence intervals across 5 folds. Maximum accuracy achieved for each variant is circled and dashed lines represent scores for Logistic Regression models trained on BOW features.}
            \label{fig:resid-accs}
        \end{figure*}

\begin{figure*}[htbp]
            \centering
            \includegraphics[width=\linewidth]{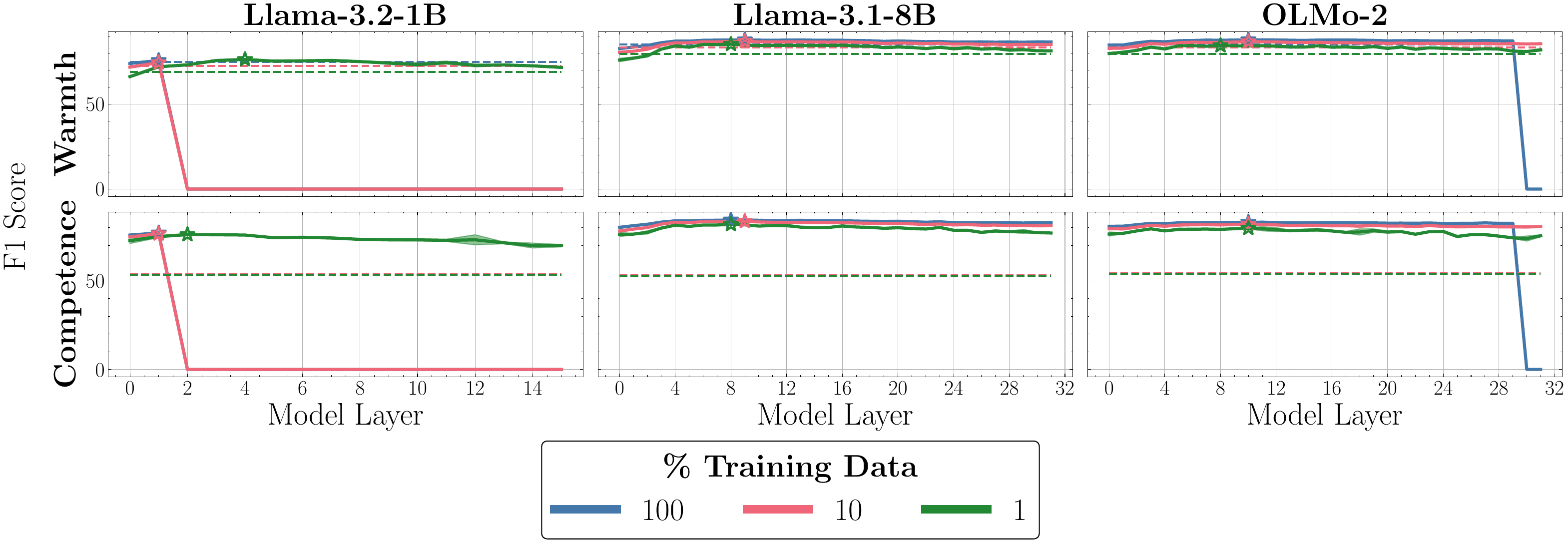}
            \caption{Accuracy scores of trained impression probes for each model, layer depth, and impression dimension. Shaded regions reflect 95\% confidence intervals across 5 folds. Maximum accuracy achieved for each variant is circled and dashed lines represent scores for Logistic Regression models trained on BOW features.}
            \label{fig:z-f1s}
        \end{figure*}

\begin{figure*}[htbp]
            \centering
            \includegraphics[width=\linewidth]{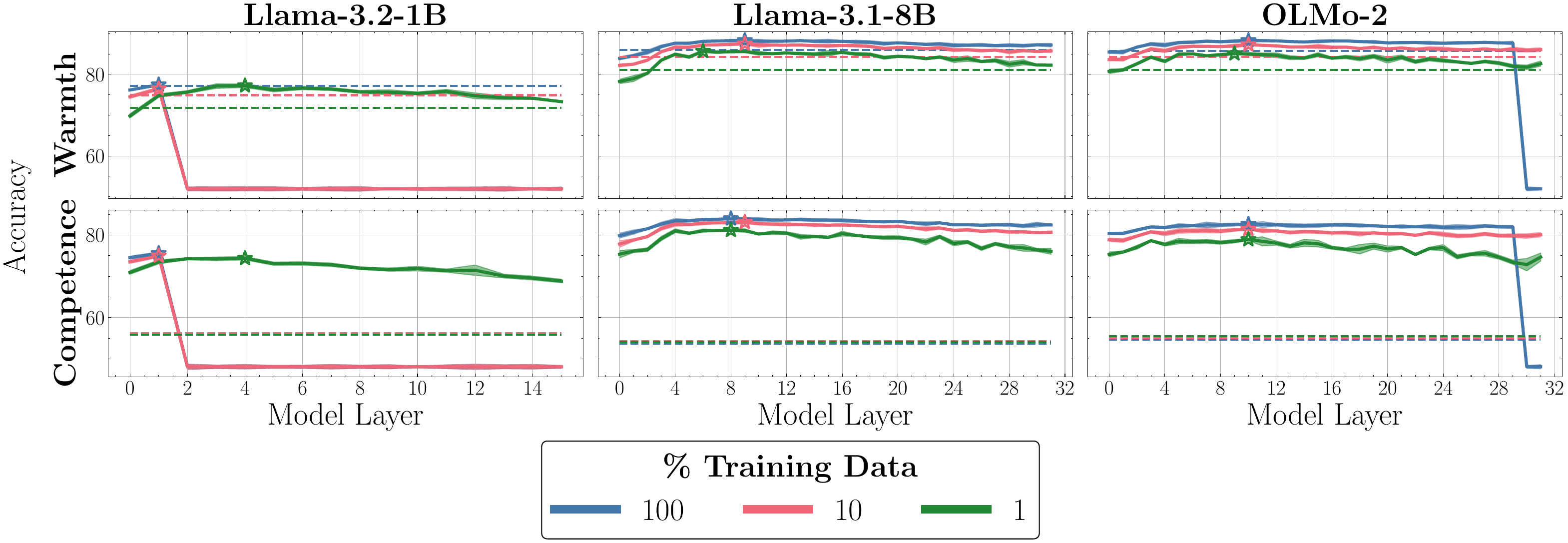}
            \caption{Accuracy scores of trained impression probes for each model, layer depth, and impression dimension. Shaded regions reflect 95\% confidence intervals across 5 folds. Maximum accuracy achieved for each variant is circled and dashed lines represent scores for Logistic Regression models trained on BOW features.}
            \label{fig:z-accs}
        \end{figure*}

        We additionally evaluate the performance of probes trained on residual streams (\autoref{fig:resid-f1s} for F1 scores and \autoref{fig:resid-accs} for accuracies) and z activations (\autoref{fig:z-f1s} for F1 scores and \autoref{fig:z-accs} for accuracies). Probes fit on residual streams roughly mimic the performance of those fit on MLP activations. Fitting on z activations for some model layers was numerically unstable, but stable probes similarly exceed the performance of the BOW baseline. Because the probes fit on MLP activations exceed or achieve similar performance to those fit on alternative representations, we focus on these probes in the main experiments. 
    
\section{Full LLM Behavior Results}
\label{app:behavior-res}

    \subsection{Response Quality}

        \autoref{tab:avg-quality} presents the average response quality for responses to LMSysChat messages. In this analysis, we notably use the binary predictions by the impression probes rather than the outputted probabilities. The average quality of responses to prompts perceived as warm and competent are higher for all models than those perceived as cold or incompetent, supporting the trend observed in \autoref{sec:rq2}. 
    
        \begin{table}[htbp]
            \centering
            \begin{tabular}{c | c c | c c }
                \hline
                \multirow{2}{*}{Model}  & \multicolumn{2}{c|}{Warmth}   & \multicolumn{2}{c}{Competence} \\
                                        & Low       & High              & Low       & High               \\
                \hline
                \llamaSmall{}           & 6.08      & \textbf{6.52}$^*$ & 5.81      & \textbf{6.32}$^*$     \\
                \llamaBig{}             & 7.72      & \textbf{7.99}     & 7.70      & \textbf{7.78}         \\
                \olmo{}                 & 7.37      & \textbf{7.81}$^*$ & 7.38      & \textbf{7.51}         \\
                \hline
            \end{tabular}
            \caption{Average quality scores among subsets of model responses to LMSysChat messages according to the discrete impression probe predictions. $^*$ indicates significant difference between subsets, and larger scores between subsets are \textbf{bolded}.}
            \label{tab:avg-quality}
        \end{table}
        
    \subsection{Hedging}

        \autoref{tab:avg-hedge} similarly presents the average count of hedge terms in model responses to LMSysChat messages. The average count of hedge terms is lower for the low competence subset than the high competence subset, supporting the trend observed in \autoref{sec:rq2}.

        \begin{table}[htbp]
            \centering
            \begin{tabular}{ c | c c | c c}
                \hline
                Model           & \multicolumn{2}{c|}{Warmth}           & \multicolumn{2}{c}{Competence} \\
                                & Low                           & High              & Low               & High \\
                \hline
                \llamaSmall{}   & 0.67 & \textbf{0.84}          & \textbf{0.93}  & 0.66 \\
                \llamaBig{}     & 0.60 & \textbf{1.03}$^{*}$    & \textbf{0.72}  & 0.62 \\
                \olmo{}         & \textbf{1.18}$^{*}$ & 0.78    & \textbf{1.13}  & 1.07 \\
                \hline
            \end{tabular}
            \caption{Average count of hedge terms among subsets of model responses to LMSysChat prompts according to discrete impression probe predictions. $*$ indicates significant difference between subsets, and larger scores between subsets are \textbf{bolded}. }
            \label{tab:avg-hedge}
        \end{table}

\section{Statistical Models and Analyses}
\label{app:stats-models}

    In our LLM response quality experiments, we choose an ordered logistic regression (or proportional-odds logistic regression) model because LLM response quality is an integer ordinal variable. Furthermore, we choose a negative binomial regression model for our hedging experiments because the number of hedge terms is a count variable. While a Poisson model would also be applicable, we primarily rely on negative binomial regression to avoid relying on the equidispersion assumption (see \autoref{app:behavior-res} for analysis of this choice).

    In our RQ3 experiment examining the impact of content and style on model responses, we use log-odds-ratios with informative Dirichlet priors \cite{monroe-idp} to conduct our analyses. The approach uses a modified form of log-odds-ratios (i.e., for a given token, $\log{\frac{f^{s1}/(1-f^{s1})}{f^{s2}/(1-f^{s2})}}$ where $f^s$ represents term frequency in subset $g$ of a corpus), using a Bayesian model with Dirichlet prior. To create an informative prior, a background corpus provides the expected distribution of terms as a background reference. In doing so, the model accounts for noise in token distributions and gives less weight to common tokens (e.g., stop words). In all experiments, we use a random sample of 10,000 LMSysChat \cite{zheng-lmsyschat} prompts as a background corpus.

\section{Warmth and Competence Comparison}
\label{app:discrim-valid}

    We analyze the relationship between warmth and competence probe predictions to understand the \textit{discriminant validity} of our approach. Discriminant validity is a type of validity defined as "the degree to which a test or measure diverges from (i.e., does not correlate with) another measure whose underlying construct is conceptually unrelated to it" \cite{vandenbos-apa}. Individual attributes may inform both warmth and competence traits; for example, casual language may be perceived as both friendly (high-warmth) and/or unprofessional (low-competence). Therefore, we expect warmth and competence measures to be minimally related or entirely distinct from each other. 
    
    \autoref{tab:warmth-comp-corr} characterizes the relationship between warmth and competence probe predictions on our subsample of LMSysChat user messages. For all models, warmth and competence are significantly negatively but weakly correlated. Additionally, warmth and competence impressions only align 38-40\% of the time. Due to the lack of a strong positive relationship, we conclude that our probing approach is capable of distinguishing between warmth and competence dimensions, supporting its discriminant validity.

    \begin{table}[htbp]
        \centering
        \begin{tabular}{ c | c c }
            \hline
            Model           & Pearson $r$ & \% Match \\
            \hline
            \llamaSmall{}   & -0.15$^*$ & 37.30\% \\
            \llamaBig{}     & -0.27$^*$ & 28.40\% \\
            \olmo{}         & -0.12$^*$ & 39.05\% \\
            \hline
        \end{tabular}
        \caption{Correlation and \% match of warmth and competence impression probe predictions on LMSysChat messages. $^*$ indicates significant correlation ($p\leq0.05$)}
        \label{tab:warmth-comp-corr}
    \end{table}

\section{IDP Analyses}
\label{app:idp-res}

    \subsection{Full LIWC Analysis: Generated Messages}

        \autoref{tab:warmth-liwc-synth} and \autoref{tab:comp-liwc-synth} present the log-odds-ratio with IDP results for warmth subsets using LIWC categories.

        \begin{table}[htbp]
            \centering
            \resizebox{\columnwidth}{!}{
                \begin{tabular}{c c c ? c c c}
                    \hline
                    \multicolumn{6}{c}{\textbf{\llamaSmall{}}} \\
                    \hline
                    \multicolumn{3}{c?}{\textbf{High Warmth}} & \multicolumn{3}{c}{\textbf{Low Warmth}} \\
                    Term & $z$ & $f$ & Term & $z$ & $f$ \\
                    \hline
                    Affiliation & \cellcolor[rgb]{0.19, 0.61, 0.32} \color{white} 125.60 & 0.8\% & Adverb & \cellcolor[rgb]{0.72, 0.08, 0.10} \color{white} -116.19 & 2.0\% \\
                    Tentative & \cellcolor[rgb]{0.55, 0.81, 0.54} \color{black} 94.10 & 1.5\% & Interrogative & \cellcolor[rgb]{0.89, 0.19, 0.15} \color{white} -103.05 & 1.7\% \\
                    Drives & \cellcolor[rgb]{0.59, 0.83, 0.57} \color{black} 90.60 & 2.0\% & Cause & \cellcolor[rgb]{0.99, 0.47, 0.35} \color{black} -84.56 & 1.1\% \\
                    Quantity & \cellcolor[rgb]{0.86, 0.95, 0.84} \color{black} 61.08 & 0.6\% & Negate & \cellcolor[rgb]{0.99, 0.56, 0.44} \color{black} -78.38 & 0.4\% \\
                    Motion & \cellcolor[rgb]{0.89, 0.96, 0.87} \color{black} 57.20 & 0.4\% & You & \cellcolor[rgb]{0.99, 0.63, 0.51} \color{black} -73.92 & 1.2\% \\
                    Conjunction & \cellcolor[rgb]{0.93, 0.97, 0.92} \color{black} 49.44 & 2.1\% & Money & \cellcolor[rgb]{0.99, 0.84, 0.77} \color{black} -58.63 & 0.2\% \\
                    Anxious & \cellcolor[rgb]{0.94, 0.98, 0.92} \color{black} 47.96 & 0.1\% & Anger & \cellcolor[rgb]{0.99, 0.84, 0.77} \color{black} -58.34 & 0.0\% \\
                    Discrepancy & \cellcolor[rgb]{0.95, 0.98, 0.94} \color{black} 45.30 & 0.7\% & Focus Present & \cellcolor[rgb]{1.00, 0.88, 0.83} \color{black} -55.02 & 4.7\% \\
                    Power & \cellcolor[rgb]{0.97, 0.99, 0.96} \color{black} 41.11 & 0.6\% & Filler & \cellcolor[rgb]{1.00, 0.89, 0.85} \color{black} -53.41 & 0.0\% \\
                    Personal Pron & \cellcolor[rgb]{0.97, 0.99, 0.96} \color{black} 40.92 & 4.9\% & Impersonal Pron & \cellcolor[rgb]{1.00, 0.89, 0.85} \color{black} -53.40 & 1.7\% \\
                    \hline
                \end{tabular}
            }
            
            \vspace{1em}
            
            \resizebox{\columnwidth}{!}{
                \begin{tabular}{c c c ? c c c}
                    \hline
                    \multicolumn{6}{c}{\textbf{\llamaBig{}}} \\
                    \hline
                    \multicolumn{3}{c?}{\textbf{High Warmth}} & \multicolumn{3}{c}{\textbf{Low Warmth}} \\
                    Term & $z$ & $f$ & Term & $z$ & $f$ \\
                    \hline
                    Affiliation & \cellcolor[rgb]{0.00, 0.27, 0.11} \color{white} 163.44 & 0.8\% & Negate & \cellcolor[rgb]{0.40, 0.00, 0.05} \color{white} -132.65 & 0.5\% \\
                    Drives & \cellcolor[rgb]{0.06, 0.48, 0.22} \color{white} 141.16 & 2.2\% & Adverb & \cellcolor[rgb]{0.84, 0.13, 0.13} \color{white} -107.13 & 1.8\% \\
                    Achieve & \cellcolor[rgb]{0.66, 0.86, 0.64} \color{black} 83.76 & 0.6\% & Impersonal Pron & \cellcolor[rgb]{0.98, 0.43, 0.30} \color{black} -87.47 & 1.4\% \\
                    Anxious & \cellcolor[rgb]{0.76, 0.90, 0.73} \color{black} 74.23 & 0.1\% & You & \cellcolor[rgb]{0.99, 0.49, 0.37} \color{black} -83.12 & 1.5\% \\
                    + Emotion & \cellcolor[rgb]{0.77, 0.91, 0.74} \color{black} 72.53 & 1.1\% & Focus Present & \cellcolor[rgb]{0.99, 0.63, 0.51} \color{black} -73.39 & 4.7\% \\
                    Motion & \cellcolor[rgb]{0.78, 0.91, 0.75} \color{black} 72.00 & 0.5\% & Interrogative & \cellcolor[rgb]{0.99, 0.72, 0.61} \color{black} -67.60 & 1.3\% \\
                    Power & \cellcolor[rgb]{0.78, 0.91, 0.75} \color{black} 71.96 & 0.8\% & Money & \cellcolor[rgb]{0.99, 0.75, 0.65} \color{black} -65.34 & 0.2\% \\
                    Quantity & \cellcolor[rgb]{0.80, 0.92, 0.77} \color{black} 69.24 & 0.8\% & Function & \cellcolor[rgb]{0.99, 0.77, 0.68} \color{black} -63.63 & 18.6\% \\
                    Affect & \cellcolor[rgb]{0.80, 0.92, 0.78} \color{black} 68.84 & 1.5\% & Certain & \cellcolor[rgb]{0.99, 0.83, 0.76} \color{black} -59.44 & 0.4\% \\
                    Tentative & \cellcolor[rgb]{0.84, 0.94, 0.81} \color{black} 63.96 & 1.5\% & Article & \cellcolor[rgb]{1.00, 0.89, 0.85} \color{black} -53.56 & 1.7\% \\
                    \hline
                \end{tabular}
            }

            \vspace{1em}

            \resizebox{\columnwidth}{!}{
                \begin{tabular}{c c c ? c c c}
                    \hline
                    \multicolumn{6}{c}{\textbf{\olmo{}}} \\
                    \hline
                    \multicolumn{3}{c?}{\textbf{High Warmth}} & \multicolumn{3}{c}{\textbf{Low Warmth}} \\
                    Term & $z$ & $f$ & Term & $z$ & $f$ \\
                    \hline
                    Affiliation & \cellcolor[rgb]{0.02, 0.45, 0.19} \color{white} 145.73 & 1.3\% & Differ & \cellcolor[rgb]{0.51, 0.03, 0.06} \color{white} -127.62 & 0.7\% \\
                    Drives & \cellcolor[rgb]{0.19, 0.61, 0.32} \color{white} 125.50 & 2.5\% & Negate & \cellcolor[rgb]{0.65, 0.06, 0.08} \color{white} -121.26 & 0.4\% \\
                    Discrepancy & \cellcolor[rgb]{0.65, 0.86, 0.62} \color{black} 85.06 & 1.1\% & Adverb & \cellcolor[rgb]{0.98, 0.38, 0.27} \color{black} -90.41 & 2.3\% \\
                    Quantity & \cellcolor[rgb]{0.78, 0.91, 0.75} \color{black} 71.77 & 0.8\% & Certain & \cellcolor[rgb]{0.99, 0.51, 0.38} \color{black} -82.16 & 0.5\% \\
                    Work & \cellcolor[rgb]{0.79, 0.92, 0.77} \color{black} 69.93 & 0.5\% & Interrogative & \cellcolor[rgb]{0.99, 0.53, 0.41} \color{black} -80.17 & 1.3\% \\
                    Power & \cellcolor[rgb]{0.81, 0.92, 0.78} \color{black} 68.32 & 0.9\% & Anger & \cellcolor[rgb]{0.99, 0.59, 0.46} \color{black} -76.69 & 0.1\% \\
                    Preposition & \cellcolor[rgb]{0.82, 0.93, 0.80} \color{black} 66.42 & 3.8\% & Impersonal Pron & \cellcolor[rgb]{0.99, 0.67, 0.56} \color{black} -70.80 & 1.0\% \\
                    + Emotion & \cellcolor[rgb]{0.90, 0.96, 0.88} \color{black} 55.46 & 1.3\% & - Emotion & \cellcolor[rgb]{0.99, 0.72, 0.61} \color{black} -67.68 & 0.4\% \\
                    Motion & \cellcolor[rgb]{0.94, 0.98, 0.93} \color{black} 46.59 & 0.5\% & Focus Present & \cellcolor[rgb]{0.99, 0.80, 0.71} \color{black} -61.78 & 4.6\% \\
                    Reward & \cellcolor[rgb]{0.94, 0.98, 0.93} \color{black} 46.28 & 0.4\% & Sad & \cellcolor[rgb]{1.00, 0.96, 0.94} \color{black} -44.25 & 0.1\% \\
                    \hline
                \end{tabular}
            }
            
            \caption{Log-odds-ratio with IDP results using LIWC categories among warmth subsets of generated prompts. $z$-scores reflect the extent that each category is associated with the high (positive) or low (negative) warmth subsets, and $f$ reflects category frequency in the full corpus.}
            \label{tab:warmth-liwc-synth}
        \end{table}

        \begin{table}[htbp]
            \centering
            \resizebox{\columnwidth}{!}{
                \begin{tabular}{c c c ? c c c}
                    \hline
                    \multicolumn{6}{c}{\textbf{\llamaBig{}}} \\
                    \hline
                    \multicolumn{3}{c?}{\textbf{High Competence}} & \multicolumn{3}{c}{\textbf{Low Competence}} \\
                    Term & $z$ & $N$ & Term & $z$ & $N$ \\
                    \hline
                    Preposition & \cellcolor[rgb]{0.67, 0.87, 0.64} \color{black} 65.93 & 4.1\% & Informal & \cellcolor[rgb]{0.40, 0.00, 0.05} \color{white} -125.16 & 1.0\% \\
                    Article & \cellcolor[rgb]{0.71, 0.89, 0.69} \color{black} 62.38 & 2.1\% & Netspeak & \cellcolor[rgb]{0.79, 0.09, 0.11} \color{white} -102.27 & 0.8\% \\
                    Adjective & \cellcolor[rgb]{0.73, 0.89, 0.71} \color{black} 61.10 & 1.1\% & Negate & \cellcolor[rgb]{0.98, 0.40, 0.28} \color{black} -79.89 & 0.4\% \\
                    Social & \cellcolor[rgb]{0.82, 0.93, 0.79} \color{black} 54.21 & 3.5\% & Death & \cellcolor[rgb]{0.99, 0.79, 0.70} \color{black} -51.96 & 0.0\% \\
                    Compare & \cellcolor[rgb]{0.89, 0.96, 0.87} \color{black} 47.04 & 0.5\% & Filler & \cellcolor[rgb]{0.99, 0.79, 0.70} \color{black} -51.81 & 0.0\% \\
                    Quantity & \cellcolor[rgb]{0.90, 0.96, 0.88} \color{black} 46.72 & 0.6\% & Focus Present & \cellcolor[rgb]{0.99, 0.82, 0.75} \color{black} -48.90 & 4.7\% \\
                    Work & \cellcolor[rgb]{0.90, 0.96, 0.88} \color{black} 46.56 & 0.6\% & Impersonal Pron & \cellcolor[rgb]{0.99, 0.85, 0.78} \color{black} -47.39 & 1.7\% \\
                    Space & \cellcolor[rgb]{0.94, 0.98, 0.93} \color{black} 40.37 & 1.7\% & Ingest & \cellcolor[rgb]{0.99, 0.85, 0.78} \color{black} -47.28 & 0.1\% \\
                    Relative & \cellcolor[rgb]{0.96, 0.98, 0.95} \color{black} 37.69 & 3.1\% & Adverb & \cellcolor[rgb]{1.00, 0.86, 0.80} \color{black} -46.17 & 2.0\% \\
                    Achieve & \cellcolor[rgb]{0.97, 0.99, 0.96} \color{black} 35.80 & 0.5\% & Interrogative & \cellcolor[rgb]{1.00, 0.89, 0.84} \color{black} -42.96 & 1.7\% \\
                    \hline
                \end{tabular}
            }

            \vspace{1em}

            \resizebox{\columnwidth}{!}{
                \begin{tabular}{c c c ? c c c}
                    \hline
                    \multicolumn{6}{c}{\textbf{\llamaSmall{}}} \\
                    \hline
                    \multicolumn{3}{c?}{\textbf{High Competence}} & \multicolumn{3}{c}{\textbf{Low Competence}} \\
                    Term & $z$ & $N$ & Term & $z$ & $N$ \\
                    \hline
                    Preposition & \cellcolor[rgb]{0.00, 0.27, 0.11} \color{white} 122.74 & 3.9\% & Adverb & \cellcolor[rgb]{0.63, 0.05, 0.08} \color{white} -114.54 & 1.8\% \\
                    Adjective & \cellcolor[rgb]{0.00, 0.43, 0.17} \color{white} 111.76 & 1.2\% & Differ & \cellcolor[rgb]{0.73, 0.08, 0.10} \color{white} -107.18 & 1.1\% \\
                    Relative & \cellcolor[rgb]{0.01, 0.44, 0.18} \color{white} 110.53 & 3.2\% & Informal & \cellcolor[rgb]{0.86, 0.16, 0.14} \color{white} -96.78 & 0.8\% \\
                    Article & \cellcolor[rgb]{0.03, 0.45, 0.19} \color{white} 109.74 & 1.7\% & Impersonal Pron & \cellcolor[rgb]{0.94, 0.23, 0.17} \color{white} -90.43 & 1.4\% \\
                    Space & \cellcolor[rgb]{0.20, 0.61, 0.32} \color{white} 95.00 & 1.6\% & Netspeak & \cellcolor[rgb]{0.99, 0.48, 0.36} \color{black} -74.09 & 0.6\% \\
                    Achieve & \cellcolor[rgb]{0.33, 0.71, 0.40} \color{white} 86.29 & 0.6\% & Negate & \cellcolor[rgb]{0.99, 0.49, 0.36} \color{black} -73.79 & 0.5\% \\
                    Drives & \cellcolor[rgb]{0.40, 0.74, 0.44} \color{black} 82.32 & 2.2\% & Interrogative & \cellcolor[rgb]{0.99, 0.51, 0.38} \color{black} -72.35 & 1.3\% \\
                    Compare & \cellcolor[rgb]{0.45, 0.77, 0.46} \color{black} 79.58 & 0.4\% & Pronoun & \cellcolor[rgb]{0.99, 0.58, 0.45} \color{black} -67.25 & 7.1\% \\
                    Work & \cellcolor[rgb]{0.46, 0.77, 0.47} \color{black} 79.10 & 0.7\% & Cognitive Proc & \cellcolor[rgb]{0.99, 0.62, 0.50} \color{black} -64.05 & 4.8\% \\
                    Quantity & \cellcolor[rgb]{0.65, 0.86, 0.63} \color{black} 66.91 & 0.8\% & - Emotion & \cellcolor[rgb]{0.99, 0.64, 0.53} \color{black} -62.55 & 0.4\% \\
                    \hline
                \end{tabular}
            }

            \vspace{1em}

            \resizebox{\columnwidth}{!}{
                \begin{tabular}{c c c ? c c c}
                    \hline
                    \multicolumn{6}{c}{\textbf{\olmo{}}} \\
                    \hline
                    \multicolumn{3}{c?}{\textbf{High Competence}} & \multicolumn{3}{c}{\textbf{Low Competence}} \\
                    Term & $z$ & $N$ & Term & $z$ & $N$ \\
                    \hline
                    Preposition & \cellcolor[rgb]{0.38, 0.73, 0.42} \color{black} 83.68 & 3.8\% & Interrogative & \cellcolor[rgb]{0.82, 0.12, 0.12} \color{white} -100.02 & 1.3\% \\
                    Compare & \cellcolor[rgb]{0.57, 0.82, 0.56} \color{black} 72.18 & 0.6\% & Informal & \cellcolor[rgb]{0.95, 0.27, 0.20} \color{white} -88.33 & 1.1\% \\
                    Space & \cellcolor[rgb]{0.63, 0.85, 0.61} \color{black} 68.72 & 1.6\% & Netspeak & \cellcolor[rgb]{0.96, 0.33, 0.23} \color{black} -84.69 & 0.7\% \\
                    Adjective & \cellcolor[rgb]{0.63, 0.85, 0.61} \color{black} 68.29 & 1.4\% & Impersonal Pron & \cellcolor[rgb]{0.98, 0.40, 0.28} \color{black} -80.12 & 1.0\% \\
                    Article & \cellcolor[rgb]{0.71, 0.89, 0.69} \color{black} 62.35 & 1.5\% & Cause & \cellcolor[rgb]{0.99, 0.47, 0.34} \color{black} -75.31 & 1.2\% \\
                    Relative & \cellcolor[rgb]{0.75, 0.90, 0.72} \color{black} 59.90 & 3.1\% & Adverb & \cellcolor[rgb]{0.99, 0.53, 0.40} \color{black} -70.83 & 2.3\% \\
                    Reward & \cellcolor[rgb]{0.90, 0.96, 0.88} \color{black} 46.57 & 0.4\% & Verb & \cellcolor[rgb]{1.00, 0.91, 0.87} \color{black} -39.88 & 4.3\% \\
                    Time & \cellcolor[rgb]{0.95, 0.98, 0.94} \color{black} 38.29 & 1.1\% & - Emotion & \cellcolor[rgb]{1.00, 0.91, 0.87} \color{black} -39.82 & 0.4\% \\
                    Insight & \cellcolor[rgb]{0.96, 0.98, 0.95} \color{black} 37.34 & 1.6\% & Affiliation & \cellcolor[rgb]{1.00, 0.95, 0.92} \color{black} -35.12 & 1.3\% \\
                    You & \cellcolor[rgb]{0.97, 0.99, 0.96} \color{black} 35.98 & 2.6\% & Feel & \cellcolor[rgb]{1.00, 0.96, 0.94} \color{black} -33.05 & 0.1\% \\
                    \hline
                \end{tabular}
            }
            \caption{Log-odds-ratio with informative Dirichlet prior results using LIWC categories among competence subsets of generated prompts. $z$-scores reflect the extent that each category is associated with the high (positive) or low (negative) warmth subsets, and $f$ reflects category frequency in the full corpus.}
            \label{tab:comp-liwc-synth}
        \end{table}

    \subsection{Full Token-Level Analysis: Synthetic Messages}

        \autoref{tab:warmth-tok-synth} and \autoref{tab:comp-tok-synth} present the log-odds-ratio with IDP results for warmth subsets using tokens.

        \begin{table}[htbp]
            \centering
            \resizebox{\columnwidth}{!}{
                \begin{tabular}{c c c ? c c c}
                    \hline
                    \multicolumn{6}{c}{\textbf{\llamaBig{}}} \\
                    \hline
                    \multicolumn{3}{c?}{\textbf{High Warmth}} & \multicolumn{3}{c}{\textbf{Low Warmth}} \\
                    Term & $z$ & $N$ & Term & $z$ & $N$ \\
                    \hline
                    im & \cellcolor[rgb]{0.18, 0.59, 0.30} \color{white} 113.82 & 1.9\% & why & \cellcolor[rgb]{0.79, 0.09, 0.11} \color{white} -116.30 & 0.5\% \\
                    hello & \cellcolor[rgb]{0.50, 0.79, 0.50} \color{black} 92.98 & 0.4\% & just & \cellcolor[rgb]{0.99, 0.71, 0.60} \color{black} -74.04 & 0.7\% \\
                    some & \cellcolor[rgb]{0.55, 0.81, 0.54} \color{black} 90.22 & 0.7\% & even & \cellcolor[rgb]{0.99, 0.80, 0.71} \color{black} -67.36 & 0.2\% \\
                    could & \cellcolor[rgb]{0.64, 0.86, 0.62} \color{black} 84.35 & 0.4\% & tell & \cellcolor[rgb]{0.99, 0.83, 0.76} \color{black} -65.09 & 0.5\% \\
                    hi & \cellcolor[rgb]{0.65, 0.86, 0.62} \color{black} 84.04 & 0.3\% & dont & \cellcolor[rgb]{0.99, 0.83, 0.76} \color{black} -65.08 & 0.3\% \\
                    wondering & \cellcolor[rgb]{0.85, 0.94, 0.83} \color{black} 68.54 & 0.2\% & is & \cellcolor[rgb]{1.00, 0.87, 0.82} \color{black} -61.77 & 1.2\% \\
                    looking & \cellcolor[rgb]{0.90, 0.96, 0.89} \color{black} 62.99 & 0.2\% & the & \cellcolor[rgb]{1.00, 0.89, 0.84} \color{black} -60.00 & 3.4\% \\
                    guidance & \cellcolor[rgb]{0.95, 0.98, 0.93} \color{black} 56.69 & 0.1\% & explain & \cellcolor[rgb]{1.00, 0.93, 0.90} \color{black} -54.28 & 0.4\% \\
                    help & \cellcolor[rgb]{0.96, 0.98, 0.95} \color{black} 55.01 & 1.0\% & cant & \cellcolor[rgb]{1.00, 0.95, 0.92} \color{black} -52.20 & 0.1\% \\
                    trouble & \cellcolor[rgb]{0.97, 0.99, 0.96} \color{black} 53.26 & 0.2\% & are & \cellcolor[rgb]{1.00, 0.96, 0.94} \color{black} -50.12 & 0.6\% \\
                    \hline
                \end{tabular}
            }

            \vspace{1em}

            \resizebox{\columnwidth}{!}{
                \begin{tabular}{c c c ? c c c}
                    \hline
                    \multicolumn{6}{c}{\textbf{\llamaSmall{}}} \\
                    \hline
                    \multicolumn{3}{c?}{\textbf{High Warmth}} & \multicolumn{3}{c}{\textbf{Low Warmth}} \\
                    Term & $z$ & $N$ & Term & $z$ & $N$ \\
                    \hline
                    im & \cellcolor[rgb]{0.00, 0.27, 0.11} \color{white} 138.49 & 2.5\% & just & \cellcolor[rgb]{0.40, 0.00, 0.05} \color{white} -138.29 & 1.1\% \\
                    some & \cellcolor[rgb]{0.00, 0.39, 0.16} \color{white} 130.29 & 1.0\% & dont & \cellcolor[rgb]{0.96, 0.30, 0.22} \color{white} -100.81 & 0.5\% \\
                    could & \cellcolor[rgb]{0.19, 0.61, 0.32} \color{white} 112.18 & 0.7\% & answer & \cellcolor[rgb]{0.99, 0.52, 0.39} \color{black} -87.12 & 0.2\% \\
                    hi & \cellcolor[rgb]{0.36, 0.72, 0.42} \color{black} 100.76 & 0.4\% & tell & \cellcolor[rgb]{0.99, 0.54, 0.41} \color{black} -85.74 & 0.5\% \\
                    help & \cellcolor[rgb]{0.58, 0.83, 0.57} \color{black} 88.28 & 1.3\% & even & \cellcolor[rgb]{0.99, 0.58, 0.45} \color{black} -82.91 & 0.3\% \\
                    hello & \cellcolor[rgb]{0.62, 0.85, 0.60} \color{black} 85.78 & 0.3\% & without & \cellcolor[rgb]{0.99, 0.62, 0.50} \color{black} -79.80 & 0.2\% \\
                    feeling & \cellcolor[rgb]{0.77, 0.91, 0.74} \color{black} 75.53 & 0.2\% & why & \cellcolor[rgb]{0.99, 0.66, 0.55} \color{black} -77.30 & 0.2\% \\
                    was & \cellcolor[rgb]{0.80, 0.92, 0.78} \color{black} 72.67 & 0.4\% & give & \cellcolor[rgb]{0.99, 0.69, 0.58} \color{black} -75.34 & 0.3\% \\
                    and & \cellcolor[rgb]{0.81, 0.93, 0.79} \color{black} 71.62 & 3.5\% & youre & \cellcolor[rgb]{0.99, 0.70, 0.60} \color{black} -74.54 & 0.3\% \\
                    wondering & \cellcolor[rgb]{0.82, 0.93, 0.79} \color{black} 71.54 & 0.4\% & the & \cellcolor[rgb]{0.99, 0.71, 0.61} \color{black} -73.72 & 2.7\% \\
                    \hline
                \end{tabular}
            }

            \vspace{1em}

            \resizebox{\columnwidth}{!}{
                \begin{tabular}{c c c ? c c c}
                    \hline
                    \multicolumn{6}{c}{\textbf{\olmo{}}} \\
                    \hline
                    \multicolumn{3}{c?}{\textbf{High Warmth}} & \multicolumn{3}{c}{\textbf{Low Warmth}} \\
                    Term & $z$ & $N$ & Term & $z$ & $N$ \\
                    \hline
                    could & \cellcolor[rgb]{0.00, 0.40, 0.16} \color{white} 129.52 & 2.5\% & are & \cellcolor[rgb]{0.74, 0.08, 0.10} \color{white} -120.70 & 1.1\% \\
                    some & \cellcolor[rgb]{0.13, 0.53, 0.26} \color{white} 118.03 & 1.1\% & just & \cellcolor[rgb]{0.84, 0.13, 0.13} \color{white} -112.98 & 0.9\% \\
                    hello & \cellcolor[rgb]{0.16, 0.57, 0.29} \color{white} 115.38 & 1.4\% & why & \cellcolor[rgb]{0.99, 0.47, 0.34} \color{black} -90.47 & 0.5\% \\
                    on & \cellcolor[rgb]{0.70, 0.88, 0.67} \color{black} 80.54 & 1.3\% & even & \cellcolor[rgb]{0.99, 0.55, 0.42} \color{black} -85.08 & 0.5\% \\
                    there & \cellcolor[rgb]{0.71, 0.88, 0.68} \color{black} 79.79 & 0.6\% & do & \cellcolor[rgb]{0.99, 0.69, 0.58} \color{black} -75.52 & 0.6\% \\
                    about & \cellcolor[rgb]{0.77, 0.91, 0.74} \color{black} 75.65 & 0.9\% & without & \cellcolor[rgb]{0.99, 0.69, 0.58} \color{black} -75.32 & 0.4\% \\
                    help & \cellcolor[rgb]{0.79, 0.92, 0.77} \color{black} 73.55 & 1.5\% & bot & \cellcolor[rgb]{0.99, 0.81, 0.74} \color{black} -66.16 & 0.8\% \\
                    how & \cellcolor[rgb]{0.85, 0.94, 0.83} \color{black} 68.29 & 1.6\% & answer & \cellcolor[rgb]{0.99, 0.82, 0.75} \color{black} -65.60 & 0.2\% \\
                    guide & \cellcolor[rgb]{0.85, 0.94, 0.83} \color{black} 68.06 & 0.5\% & all & \cellcolor[rgb]{1.00, 0.86, 0.80} \color{black} -62.55 & 0.3\% \\
                    assist & \cellcolor[rgb]{0.86, 0.94, 0.83} \color{black} 67.83 & 0.7\% & or & \cellcolor[rgb]{1.00, 0.88, 0.83} \color{black} -60.60 & 0.5\% \\
                    \hline
                \end{tabular}
            }
            \caption{Log-odds-ratio with IDP results using tokens among warmth subsets of generated prompts. $z$-scores reflect the extent that each category is associated with the high (positive) or low (negative) warmth subsets, and $f$ reflects term frequency in the full corpus.}
            \label{tab:warmth-tok-synth}
        \end{table}

        \begin{table}[htbp]
            \centering
            \resizebox{\columnwidth}{!}{
                \begin{tabular}{c c c ? c c c}
                    \hline
                    \multicolumn{6}{c}{\textbf{\llamaBig{}}} \\
                    \hline
                    \multicolumn{3}{c?}{\textbf{High Competence}} & \multicolumn{3}{c}{\textbf{Low Competence}} \\
                    Term & $z$ & $N$ & Term & $z$ & $N$ \\
                    \hline
                    provide & \cellcolor[rgb]{0.65, 0.86, 0.63} \color{black} 56.69 & 0.3\% & dont & \cellcolor[rgb]{0.91, 0.20, 0.16} \color{white} -92.30 & 0.3\% \\
                    your & \cellcolor[rgb]{0.83, 0.94, 0.81} \color{black} 45.62 & 0.5\% & just & \cellcolor[rgb]{0.96, 0.31, 0.23} \color{white} -86.06 & 0.7\% \\
                    looking & \cellcolor[rgb]{0.90, 0.96, 0.88} \color{black} 40.96 & 0.2\% & really & \cellcolor[rgb]{0.99, 0.52, 0.39} \color{black} -75.09 & 0.3\% \\
                    information & \cellcolor[rgb]{0.90, 0.96, 0.88} \color{black} 40.88 & 0.2\% & do & \cellcolor[rgb]{1.00, 0.88, 0.82} \color{black} -53.18 & 0.8\% \\
                    of & \cellcolor[rgb]{0.90, 0.96, 0.88} \color{black} 40.71 & 1.8\% & no & \cellcolor[rgb]{1.00, 0.90, 0.85} \color{black} -51.32 & 0.1\% \\
                    concept & \cellcolor[rgb]{0.90, 0.96, 0.89} \color{black} 40.33 & 0.2\% & ur & \cellcolor[rgb]{1.00, 0.90, 0.85} \color{black} -50.84 & 0.1\% \\
                    id & \cellcolor[rgb]{0.94, 0.98, 0.92} \color{black} 36.63 & 0.1\% & know & \cellcolor[rgb]{1.00, 0.91, 0.87} \color{black} -49.42 & 0.4\% \\
                    hello & \cellcolor[rgb]{0.94, 0.98, 0.93} \color{black} 35.77 & 0.4\% & get & \cellcolor[rgb]{1.00, 0.94, 0.91} \color{black} -46.51 & 0.5\% \\
                    guidance & \cellcolor[rgb]{0.95, 0.98, 0.93} \color{black} 35.31 & 0.1\% & hey & \cellcolor[rgb]{1.00, 0.96, 0.94} \color{black} -43.95 & 0.4\% \\
                    ai & \cellcolor[rgb]{0.97, 0.99, 0.96} \color{black} 32.51 & 0.1\% & help & \cellcolor[rgb]{1.00, 0.96, 0.94} \color{black} -43.79 & 1.0\% \\
                    \hline
                \end{tabular}
            }

            \vspace{1em}

            \resizebox{\columnwidth}{!}{
                \begin{tabular}{c c c ? c c c}
                    \hline
                    \multicolumn{6}{c}{\textbf{\llamaSmall{}}} \\
                    \hline
                    \multicolumn{3}{c?}{\textbf{High Competence}} & \multicolumn{3}{c}{\textbf{Low Competence}} \\
                    Term & $z$ & $N$ & Term & $z$ & $N$ \\
                    \hline
                    provide & \cellcolor[rgb]{0.11, 0.52, 0.25} \color{white} 85.33 & 0.4\% & just & \cellcolor[rgb]{0.40, 0.00, 0.05} \color{white} -118.37 & 1.1\% \\
                    your & \cellcolor[rgb]{0.51, 0.79, 0.51} \color{black} 63.95 & 0.5\% & really & \cellcolor[rgb]{0.56, 0.04, 0.07} \color{white} -112.22 & 0.6\% \\
                    most & \cellcolor[rgb]{0.52, 0.80, 0.52} \color{black} 63.45 & 0.2\% & me & \cellcolor[rgb]{0.84, 0.13, 0.13} \color{white} -96.92 & 2.9\% \\
                    looking & \cellcolor[rgb]{0.63, 0.85, 0.61} \color{black} 58.02 & 0.2\% & if & \cellcolor[rgb]{0.91, 0.20, 0.16} \color{white} -92.34 & 0.8\% \\
                    planning & \cellcolor[rgb]{0.66, 0.86, 0.63} \color{black} 56.46 & 0.1\% & do & \cellcolor[rgb]{0.99, 0.46, 0.33} \color{black} -78.71 & 0.6\% \\
                    reaching & \cellcolor[rgb]{0.71, 0.88, 0.68} \color{black} 53.69 & 0.1\% & dont & \cellcolor[rgb]{0.99, 0.47, 0.35} \color{black} -77.77 & 0.5\% \\
                    trip & \cellcolor[rgb]{0.71, 0.88, 0.68} \color{black} 53.66 & 0.1\% & tell & \cellcolor[rgb]{0.99, 0.51, 0.38} \color{black} -75.79 & 0.5\% \\
                    excited & \cellcolor[rgb]{0.71, 0.89, 0.69} \color{black} 53.34 & 0.1\% & know & \cellcolor[rgb]{0.99, 0.51, 0.39} \color{black} -75.35 & 0.4\% \\
                    on & \cellcolor[rgb]{0.74, 0.90, 0.72} \color{black} 51.75 & 1.0\% & but & \cellcolor[rgb]{0.99, 0.57, 0.45} \color{black} -72.02 & 0.6\% \\
                    in & \cellcolor[rgb]{0.74, 0.90, 0.72} \color{black} 51.66 & 0.9\% & sure & \cellcolor[rgb]{0.99, 0.57, 0.45} \color{black} -71.78 & 0.3\% \\
                    \hline
                \end{tabular}
            }

            \vspace{1em}

            \resizebox{\columnwidth}{!}{
                \begin{tabular}{c c c ? c c c}
                    \hline
                    \multicolumn{6}{c}{\textbf{\olmo{}}} \\
                    \hline
                    \multicolumn{3}{c?}{\textbf{High Competence}} & \multicolumn{3}{c}{\textbf{Low Competence}} \\
                    Term & $z$ & $N$ & Term & $z$ & $N$ \\
                    \hline
                    provide & \cellcolor[rgb]{0.00, 0.27, 0.11} \color{white} 100.48 & 1.1\% & help & \cellcolor[rgb]{0.94, 0.25, 0.18} \color{white} -89.70 & 1.5\% \\
                    your & \cellcolor[rgb]{0.53, 0.80, 0.52} \color{black} 62.96 & 0.7\% & do & \cellcolor[rgb]{0.99, 0.65, 0.53} \color{black} -67.54 & 0.6\% \\
                    on & \cellcolor[rgb]{0.56, 0.82, 0.55} \color{black} 61.19 & 1.3\% & hi & \cellcolor[rgb]{0.99, 0.68, 0.57} \color{black} -65.75 & 0.7\% \\
                    latest & \cellcolor[rgb]{0.67, 0.87, 0.64} \color{black} 55.96 & 0.3\% & me & \cellcolor[rgb]{0.99, 0.71, 0.60} \color{black} -64.03 & 4.3\% \\
                    energy & \cellcolor[rgb]{0.86, 0.94, 0.83} \color{black} 44.12 & 0.2\% & tell & \cellcolor[rgb]{0.99, 0.71, 0.61} \color{black} -63.87 & 0.6\% \\
                    of & \cellcolor[rgb]{0.86, 0.95, 0.84} \color{black} 43.59 & 1.4\% & this & \cellcolor[rgb]{0.99, 0.71, 0.61} \color{black} -63.72 & 0.7\% \\
                    detailed & \cellcolor[rgb]{0.87, 0.95, 0.85} \color{black} 43.33 & 0.2\% & how & \cellcolor[rgb]{0.99, 0.81, 0.74} \color{black} -57.37 & 1.6\% \\
                    the & \cellcolor[rgb]{0.87, 0.95, 0.85} \color{black} 43.24 & 2.8\% & please & \cellcolor[rgb]{1.00, 0.88, 0.83} \color{black} -52.66 & 0.9\% \\
                    in & \cellcolor[rgb]{0.89, 0.96, 0.87} \color{black} 41.73 & 0.7\% & what & \cellcolor[rgb]{1.00, 0.89, 0.83} \color{black} -52.28 & 0.4\% \\
                    renewable & \cellcolor[rgb]{0.90, 0.96, 0.88} \color{black} 40.63 & 0.1\% & something & \cellcolor[rgb]{1.00, 0.91, 0.86} \color{black} -50.12 & 0.3\% \\
                    \hline
                \end{tabular}
            }
            \caption{Log-odds-ratio with IDP results using tokens among competence subsets of generated prompts. $z$-scores reflect the extent that each category is associated with the high (positive) or low (negative) warmth subsets, and $f$ reflects term frequency in the full corpus.}
            \label{tab:comp-tok-synth}
        \end{table}

    \subsection{Full LIWC Analysis: LMSysChat}

        We repeat our analyses on LMSysChat \cite{zheng-lmsyschat} data to identify patterns in real world texts. \autoref{tab:warmth-tok-lmsys} and \autoref{tab:comp-tok-lmsys} present the log-odds-ratio with IDP results for warmth subsets using tokens.

        \begin{table}[htbp]
            \centering
            \resizebox{\columnwidth}{!}{
                \begin{tabular}{c c c ? c c c}
                    \hline
                    \multicolumn{6}{c}{\textbf{\llamaBig{}}} \\
                    \hline
                    \multicolumn{3}{c?}{\textbf{High Warmth}} & \multicolumn{3}{c}{\textbf{Low Warmth}} \\
                    Term & $z$ & $N$ & Term & $z$ & $N$ \\
                    \hline
                    Function & \cellcolor[rgb]{0.51, 0.79, 0.51} \color{black} 7.25 & 19.0\% & Work & \cellcolor[rgb]{0.40, 0.00, 0.05} \color{white} -2.05 & 1.5\% \\
                    Interrogative & \cellcolor[rgb]{0.87, 0.95, 0.85} \color{black} 3.15 & 1.5\% & Negate & \cellcolor[rgb]{0.95, 0.27, 0.20} \color{white} -1.35 & 0.3\% \\
                    Netspeak & \cellcolor[rgb]{0.88, 0.95, 0.86} \color{black} 2.99 & 0.6\% & Cause & \cellcolor[rgb]{0.95, 0.29, 0.21} \color{white} -1.33 & 1.3\% \\
                    Informal & \cellcolor[rgb]{0.90, 0.96, 0.88} \color{black} 2.70 & 0.7\% & Article & \cellcolor[rgb]{0.98, 0.40, 0.28} \color{black} -1.21 & 3.8\% \\
                    Pronoun & \cellcolor[rgb]{0.92, 0.97, 0.91} \color{black} 2.16 & 4.0\% & Death & \cellcolor[rgb]{0.99, 0.54, 0.41} \color{black} -1.02 & 0.0\% \\
                    Personal Pron & \cellcolor[rgb]{0.94, 0.98, 0.92} \color{black} 1.86 & 2.3\% & Sexual & \cellcolor[rgb]{0.99, 0.55, 0.43} \color{black} -1.00 & 0.1\% \\
                    Aux Verb & \cellcolor[rgb]{0.94, 0.98, 0.93} \color{black} 1.77 & 2.7\% & Certain & \cellcolor[rgb]{0.99, 0.75, 0.66} \color{black} -0.72 & 0.4\% \\
                    Home & \cellcolor[rgb]{0.96, 0.98, 0.95} \color{black} 1.39 & 0.1\% & Swear & \cellcolor[rgb]{0.99, 0.76, 0.67} \color{black} -0.71 & 0.0\% \\
                    Relative & \cellcolor[rgb]{0.96, 0.98, 0.95} \color{black} 1.34 & 4.6\% & Money & \cellcolor[rgb]{0.99, 0.77, 0.68} \color{black} -0.70 & 0.3\% \\
                    Ingest & \cellcolor[rgb]{0.97, 0.99, 0.96} \color{black} 1.18 & 0.2\% & Family & \cellcolor[rgb]{1.00, 0.86, 0.80} \color{black} -0.56 & 0.1\% \\
                    \hline
                \end{tabular}
            }

            \vspace{1em}

            \resizebox{\columnwidth}{!}{
                \begin{tabular}{c c c ? c c c}
                    \hline
                    \multicolumn{6}{c}{\textbf{\llamaSmall{}}} \\
                    \hline
                    \multicolumn{3}{c?}{\textbf{High Warmth}} & \multicolumn{3}{c}{\textbf{Low Warmth}} \\
                    Term & $z$ & $N$ & Term & $z$ & $N$ \\
                    \hline
                    Function & \cellcolor[rgb]{0.00, 0.27, 0.11} \color{white} 14.44 & 19.0\% & Female & \cellcolor[rgb]{0.95, 0.27, 0.20} \color{white} -1.35 & 0.3\% \\
                    Pronoun & \cellcolor[rgb]{0.91, 0.96, 0.89} \color{black} 2.54 & 4.0\% & Negate & \cellcolor[rgb]{0.98, 0.41, 0.28} \color{black} -1.20 & 0.3\% \\
                    Personal Pron & \cellcolor[rgb]{0.91, 0.97, 0.89} \color{black} 2.47 & 2.3\% & Focus Past & \cellcolor[rgb]{0.99, 0.61, 0.49} \color{black} -0.92 & 0.6\% \\
                    Social & \cellcolor[rgb]{0.93, 0.97, 0.91} \color{black} 2.04 & 3.3\% & Differ & \cellcolor[rgb]{0.99, 0.62, 0.50} \color{black} -0.91 & 0.8\% \\
                    Verb & \cellcolor[rgb]{0.93, 0.97, 0.91} \color{black} 2.03 & 5.3\% & Feel & \cellcolor[rgb]{0.99, 0.68, 0.57} \color{black} -0.84 & 0.2\% \\
                    Focus Present & \cellcolor[rgb]{0.94, 0.98, 0.93} \color{black} 1.73 & 4.1\% & Cause & \cellcolor[rgb]{0.99, 0.76, 0.67} \color{black} -0.72 & 1.3\% \\
                    You & \cellcolor[rgb]{0.95, 0.98, 0.94} \color{black} 1.54 & 0.7\% & Certain & \cellcolor[rgb]{0.99, 0.81, 0.73} \color{black} -0.64 & 0.4\% \\
                    Aux Verb & \cellcolor[rgb]{0.96, 0.99, 0.95} \color{black} 1.26 & 2.7\% & Death & \cellcolor[rgb]{0.99, 0.83, 0.76} \color{black} -0.62 & 0.0\% \\
                    Drives & \cellcolor[rgb]{0.97, 0.99, 0.96} \color{black} 1.14 & 2.1\% & Swear & \cellcolor[rgb]{1.00, 0.88, 0.83} \color{black} -0.53 & 0.0\% \\
                    Relative & \cellcolor[rgb]{0.97, 0.99, 0.96} \color{black} 1.08 & 4.6\% & Body & \cellcolor[rgb]{1.00, 0.96, 0.94} \color{black} -0.33 & 0.2\% \\
                    \hline
                \end{tabular}
            }

            \vspace{1em}

            \resizebox{\columnwidth}{!}{
                \begin{tabular}{c c c ? c c c}
                    \hline
                    \multicolumn{6}{c}{\textbf{\olmo{}}} \\
                    \hline
                    \multicolumn{3}{c?}{\textbf{High Warmth}} & \multicolumn{3}{c}{\textbf{Low Warmth}} \\
                    Term & $z$ & $N$ & Term & $z$ & $N$ \\
                    \hline
                    Function & \cellcolor[rgb]{0.09, 0.51, 0.24} \color{white} 11.64 & 19.0\% & Female & \cellcolor[rgb]{0.94, 0.23, 0.17} \color{white} -1.40 & 0.3\% \\
                    Social & \cellcolor[rgb]{0.94, 0.98, 0.92} \color{black} 1.86 & 3.3\% & Feel & \cellcolor[rgb]{0.96, 0.31, 0.22} \color{white} -1.31 & 0.2\% \\
                    Personal Pron & \cellcolor[rgb]{0.94, 0.98, 0.93} \color{black} 1.70 & 2.3\% & Differ & \cellcolor[rgb]{0.97, 0.35, 0.25} \color{black} -1.26 & 0.8\% \\
                    Preposition & \cellcolor[rgb]{0.94, 0.98, 0.93} \color{black} 1.68 & 5.5\% & Body & \cellcolor[rgb]{0.99, 0.45, 0.33} \color{black} -1.14 & 0.2\% \\
                    Verb & \cellcolor[rgb]{0.95, 0.98, 0.94} \color{black} 1.59 & 5.3\% & Ingest & \cellcolor[rgb]{0.99, 0.47, 0.35} \color{black} -1.11 & 0.2\% \\
                    Pronoun & \cellcolor[rgb]{0.95, 0.98, 0.94} \color{black} 1.57 & 4.0\% & She/He & \cellcolor[rgb]{0.99, 0.55, 0.42} \color{black} -1.01 & 0.2\% \\
                    You & \cellcolor[rgb]{0.96, 0.98, 0.95} \color{black} 1.44 & 0.7\% & Money & \cellcolor[rgb]{0.99, 0.56, 0.44} \color{black} -0.99 & 0.3\% \\
                    Focus Present & \cellcolor[rgb]{0.96, 0.98, 0.95} \color{black} 1.37 & 4.1\% & Death & \cellcolor[rgb]{0.99, 0.81, 0.74} \color{black} -0.64 & 0.0\% \\
                    Affiliation & \cellcolor[rgb]{0.96, 0.98, 0.95} \color{black} 1.31 & 0.6\% & Negate & \cellcolor[rgb]{0.99, 0.82, 0.74} \color{black} -0.63 & 0.3\% \\
                    Drives & \cellcolor[rgb]{0.97, 0.99, 0.96} \color{black} 1.13 & 2.1\% & Swear & \cellcolor[rgb]{0.99, 0.82, 0.74} \color{black} -0.63 & 0.0\% \\
                    \hline
                \end{tabular}
            }
            \caption{Log-odds-ratio with IDP results using LIWC categories among warmth subsets of LMSysChat prompts. $z$-scores reflect the extent that each category is associated with the high (positive) or low (negative) warmth subsets, and $f$ reflects category frequency in the full corpus.}
            \label{tab:warmth-liwc-lmsys}
        \end{table}

        \begin{table}[htbp]
            \centering
            \resizebox{\columnwidth}{!}{
                \begin{tabular}{c c c ? c c c}
                    \hline
                    \multicolumn{6}{c}{\textbf{\llamaBig{}}} \\
                    \hline
                    \multicolumn{3}{c?}{\textbf{High Competence}} & \multicolumn{3}{c}{\textbf{Low Competence}} \\
                    Term & $z$ & $N$ & Term & $z$ & $N$ \\
                    \hline
                    Compare & \cellcolor[rgb]{0.62, 0.84, 0.60} \color{black} 0.86 & 0.8\% & Function & \cellcolor[rgb]{0.40, 0.00, 0.05} \color{white} -12.47 & 19.0\% \\
                    We & \cellcolor[rgb]{0.70, 0.88, 0.67} \color{black} 0.77 & 0.2\% & Female & \cellcolor[rgb]{1.00, 0.88, 0.82} \color{black} -2.56 & 0.3\% \\
                    Money & \cellcolor[rgb]{0.77, 0.91, 0.74} \color{black} 0.69 & 0.3\% & Pronoun & \cellcolor[rgb]{1.00, 0.90, 0.86} \color{black} -2.12 & 4.0\% \\
                    Anxious & \cellcolor[rgb]{0.84, 0.94, 0.81} \color{black} 0.60 & 0.1\% & Social & \cellcolor[rgb]{1.00, 0.91, 0.86} \color{black} -2.07 & 3.3\% \\
                    Health & \cellcolor[rgb]{0.85, 0.94, 0.83} \color{black} 0.58 & 1.4\% & Interrogative & \cellcolor[rgb]{1.00, 0.91, 0.87} \color{black} -1.99 & 1.5\% \\
                    Certain & \cellcolor[rgb]{0.91, 0.97, 0.90} \color{black} 0.48 & 0.4\% & Verb & \cellcolor[rgb]{1.00, 0.91, 0.87} \color{black} -1.94 & 5.3\% \\
                    Focus Future & \cellcolor[rgb]{0.92, 0.97, 0.91} \color{black} 0.46 & 0.3\% & Focus Present & \cellcolor[rgb]{1.00, 0.92, 0.88} \color{black} -1.82 & 4.1\% \\
                    Differ & \cellcolor[rgb]{0.94, 0.98, 0.93} \color{black} 0.42 & 0.8\% & Personal Pron & \cellcolor[rgb]{1.00, 0.93, 0.90} \color{black} -1.60 & 2.3\% \\
                    Work & \cellcolor[rgb]{0.96, 0.98, 0.95} \color{black} 0.38 & 1.5\% & Relative & \cellcolor[rgb]{1.00, 0.95, 0.92} \color{black} -1.39 & 4.6\% \\
                    Achieve & \cellcolor[rgb]{0.97, 0.99, 0.96} \color{black} 0.35 & 0.6\% & Informal & \cellcolor[rgb]{1.00, 0.96, 0.94} \color{black} -1.10 & 0.7\% \\
                    \hline
                \end{tabular}
            }

            \vspace{1em}

            \resizebox{\columnwidth}{!}{
                \begin{tabular}{c c c ? c c c}
                    \hline
                    \multicolumn{6}{c}{\textbf{\llamaSmall{}}} \\
                    \hline
                    \multicolumn{3}{c?}{\textbf{High Competence}} & \multicolumn{3}{c}{\textbf{Low Competence}} \\
                    Term & $z$ & $N$ & Term & $z$ & $N$ \\
                    \hline
                    Compare & \cellcolor[rgb]{0.06, 0.48, 0.22} \color{white} 1.43 & 0.8\% & Function & \cellcolor[rgb]{0.95, 0.27, 0.20} \color{white} -7.85 & 19.0\% \\
                    Achieve & \cellcolor[rgb]{0.50, 0.79, 0.50} \color{black} 0.97 & 0.6\% & Pronoun & \cellcolor[rgb]{1.00, 0.90, 0.85} \color{black} -2.21 & 4.0\% \\
                    Work & \cellcolor[rgb]{0.55, 0.81, 0.54} \color{black} 0.92 & 1.5\% & Personal Pron & \cellcolor[rgb]{1.00, 0.90, 0.85} \color{black} -2.21 & 2.3\% \\
                    Space & \cellcolor[rgb]{0.73, 0.89, 0.71} \color{black} 0.73 & 2.7\% & Focus Present & \cellcolor[rgb]{1.00, 0.92, 0.88} \color{black} -1.90 & 4.1\% \\
                    + Emotion & \cellcolor[rgb]{0.79, 0.92, 0.77} \color{black} 0.66 & 1.3\% & Body & \cellcolor[rgb]{1.00, 0.92, 0.89} \color{black} -1.77 & 0.2\% \\
                    Motion & \cellcolor[rgb]{0.80, 0.92, 0.78} \color{black} 0.65 & 0.7\% & Interrogative & \cellcolor[rgb]{1.00, 0.92, 0.89} \color{black} -1.74 & 1.5\% \\
                    Negate & \cellcolor[rgb]{0.84, 0.94, 0.81} \color{black} 0.60 & 0.3\% & Social & \cellcolor[rgb]{1.00, 0.93, 0.89} \color{black} -1.72 & 3.3\% \\
                    Article & \cellcolor[rgb]{0.84, 0.94, 0.81} \color{black} 0.60 & 3.8\% & Male & \cellcolor[rgb]{1.00, 0.94, 0.91} \color{black} -1.53 & 0.2\% \\
                    Focus Past & \cellcolor[rgb]{0.88, 0.95, 0.85} \color{black} 0.55 & 0.6\% & Female & \cellcolor[rgb]{1.00, 0.94, 0.91} \color{black} -1.47 & 0.3\% \\
                    Money & \cellcolor[rgb]{0.88, 0.95, 0.86} \color{black} 0.54 & 0.3\% & Verb & \cellcolor[rgb]{1.00, 0.94, 0.91} \color{black} -1.47 & 5.3\% \\
                    \hline
                \end{tabular}
            }

            \vspace{1em}

            \resizebox{\columnwidth}{!}{
                \begin{tabular}{c c c ? c c c}
                    \hline
                    \multicolumn{6}{c}{\textbf{\olmo{}}} \\
                    \hline
                    \multicolumn{3}{c?}{\textbf{High Competence}} & \multicolumn{3}{c}{\textbf{Low Competence}} \\
                    Term & $z$ & $N$ & Term & $z$ & $N$ \\
                    \hline
                    Work & \cellcolor[rgb]{0.00, 0.27, 0.11} \color{white} 1.67 & 1.5\% & Function & \cellcolor[rgb]{0.99, 0.53, 0.40} \color{black} -5.79 & 19.0\% \\
                    Compare & \cellcolor[rgb]{0.16, 0.57, 0.29} \color{white} 1.31 & 0.8\% & Female & \cellcolor[rgb]{0.99, 0.75, 0.65} \color{black} -3.81 & 0.3\% \\
                    Cause & \cellcolor[rgb]{0.40, 0.74, 0.44} \color{black} 1.05 & 1.3\% & Personal Pron & \cellcolor[rgb]{1.00, 0.87, 0.81} \color{black} -2.64 & 2.3\% \\
                    Article & \cellcolor[rgb]{0.47, 0.78, 0.47} \color{black} 1.00 & 3.8\% & Pronoun & \cellcolor[rgb]{1.00, 0.88, 0.83} \color{black} -2.44 & 4.0\% \\
                    Preposition & \cellcolor[rgb]{0.59, 0.83, 0.57} \color{black} 0.88 & 5.5\% & She/He & \cellcolor[rgb]{1.00, 0.91, 0.87} \color{black} -2.00 & 0.2\% \\
                    Achieve & \cellcolor[rgb]{0.61, 0.84, 0.59} \color{black} 0.87 & 0.6\% & Social & \cellcolor[rgb]{1.00, 0.92, 0.88} \color{black} -1.84 & 3.3\% \\
                    Anxious & \cellcolor[rgb]{0.70, 0.88, 0.68} \color{black} 0.76 & 0.1\% & Informal & \cellcolor[rgb]{1.00, 0.93, 0.90} \color{black} -1.56 & 0.7\% \\
                    Conjunction & \cellcolor[rgb]{0.78, 0.91, 0.75} \color{black} 0.68 & 2.3\% & Aux Verb & \cellcolor[rgb]{1.00, 0.94, 0.91} \color{black} -1.46 & 2.7\% \\
                    Cognitive Proc & \cellcolor[rgb]{0.81, 0.92, 0.78} \color{black} 0.64 & 4.2\% & Body & \cellcolor[rgb]{1.00, 0.95, 0.93} \color{black} -1.24 & 0.2\% \\
                    Insight & \cellcolor[rgb]{0.85, 0.94, 0.82} \color{black} 0.59 & 0.7\% & Time & \cellcolor[rgb]{1.00, 0.96, 0.94} \color{black} -1.17 & 1.2\% \\
                    \hline
                \end{tabular}
            }
            \caption{Log-odds-ratio with IDP results using LIWC categories among competence subsets of LMSysChat prompts. $z$-scores reflect the extent that each category is associated with the high (positive) or low (negative) warmth subsets, and $f$ reflects category frequency in the full corpus.}
            \label{tab:comp-liwc-lmsys}
        \end{table}

    \subsection{Full Token Analysis: LMSysChat}

        \autoref{tab:warmth-tok-lmsys} and \autoref{tab:comp-tok-lmsys} present the log-odds-ratio with IDP results for warmth subsets using tokens.

        \begin{table}[htbp]
            \centering
            \resizebox{\columnwidth}{!}{
                \begin{tabular}{c c c ? c c c}
                    \hline
                    \multicolumn{6}{c}{\textbf{\llamaBig{}}} \\
                    \hline
                    \multicolumn{3}{c?}{\textbf{High Warmth}} & \multicolumn{3}{c}{\textbf{Low Warmth}} \\
                    Term & $z$ & $N$ & Term & $z$ & $N$ \\
                    \hline
                    what & \cellcolor[rgb]{0.00, 0.27, 0.11} \color{white} 3.62 & 20.0\% & n1 & \cellcolor[rgb]{0.94, 0.25, 0.18} \color{white} -1.87 & 0.4\% \\
                    microwave & \cellcolor[rgb]{0.64, 0.86, 0.62} \color{black} 2.20 & 0.4\% & mbr & \cellcolor[rgb]{0.99, 0.48, 0.36} \color{black} -1.73 & 0.3\% \\
                    can & \cellcolor[rgb]{0.71, 0.88, 0.68} \color{black} 2.08 & 17.3\% & risk & \cellcolor[rgb]{0.99, 0.63, 0.51} \color{black} -1.64 & 1.2\% \\
                    bladder & \cellcolor[rgb]{0.79, 0.92, 0.76} \color{black} 1.93 & 0.4\% & write & \cellcolor[rgb]{1.00, 0.88, 0.83} \color{black} -1.46 & 17.2\% \\
                    integrity & \cellcolor[rgb]{0.80, 0.92, 0.77} \color{black} 1.91 & 0.3\% & bond & \cellcolor[rgb]{1.00, 0.90, 0.86} \color{black} -1.44 & 0.4\% \\
                    do & \cellcolor[rgb]{0.90, 0.96, 0.88} \color{black} 1.66 & 10.4\% & skirt & \cellcolor[rgb]{1.00, 0.90, 0.86} \color{black} -1.44 & 0.2\% \\
                    antenna & \cellcolor[rgb]{0.92, 0.97, 0.90} \color{black} 1.59 & 0.2\% & citations & \cellcolor[rgb]{1.00, 0.90, 0.86} \color{black} -1.44 & 0.2\% \\
                    tube & \cellcolor[rgb]{0.92, 0.97, 0.90} \color{black} 1.58 & 0.3\% & cameras & \cellcolor[rgb]{1.00, 0.95, 0.92} \color{black} -1.39 & 0.3\% \\
                    popcorn & \cellcolor[rgb]{0.93, 0.97, 0.91} \color{black} 1.56 & 0.2\% & wavelength & \cellcolor[rgb]{1.00, 0.95, 0.93} \color{black} -1.38 & 0.2\% \\
                    conda & \cellcolor[rgb]{0.93, 0.97, 0.91} \color{black} 1.56 & 0.3\% & saving & \cellcolor[rgb]{1.00, 0.95, 0.93} \color{black} -1.38 & 0.2\% \\
                    \hline
                \end{tabular}
            }

            \vspace{1em}

            \resizebox{\columnwidth}{!}{
                \begin{tabular}{c c c ? c c c}
                    \hline
                    \multicolumn{6}{c}{\textbf{\llamaSmall{}}} \\
                    \hline
                    \multicolumn{3}{c?}{\textbf{High Warmth}} & \multicolumn{3}{c}{\textbf{Low Warmth}} \\
                    Term & $z$ & $N$ & Term & $z$ & $N$ \\
                    \hline
                    you & \cellcolor[rgb]{0.46, 0.77, 0.47} \color{black} 2.50 & 32.0\% & microwave & \cellcolor[rgb]{0.94, 0.25, 0.19} \color{white} -1.87 & 0.4\% \\
                    the & \cellcolor[rgb]{0.65, 0.86, 0.62} \color{black} 2.19 & 122.0\% & bond & \cellcolor[rgb]{0.97, 0.36, 0.26} \color{black} -1.81 & 0.4\% \\
                    to & \cellcolor[rgb]{0.67, 0.87, 0.64} \color{black} 2.16 & 80.8\% & n1 & \cellcolor[rgb]{0.97, 0.36, 0.26} \color{black} -1.81 & 0.4\% \\
                    and & \cellcolor[rgb]{0.89, 0.96, 0.87} \color{black} 1.69 & 65.9\% & mbr & \cellcolor[rgb]{0.99, 0.58, 0.46} \color{black} -1.67 & 0.3\% \\
                    my & \cellcolor[rgb]{0.91, 0.96, 0.89} \color{black} 1.62 & 9.8\% & lights & \cellcolor[rgb]{0.99, 0.75, 0.65} \color{black} -1.56 & 0.4\% \\
                    accessory & \cellcolor[rgb]{0.91, 0.97, 0.89} \color{black} 1.62 & 0.2\% & risk & \cellcolor[rgb]{0.99, 0.85, 0.79} \color{black} -1.49 & 1.2\% \\
                    river & \cellcolor[rgb]{0.95, 0.98, 0.94} \color{black} 1.47 & 0.4\% & brand & \cellcolor[rgb]{1.00, 0.91, 0.87} \color{black} -1.43 & 1.1\% \\
                    are & \cellcolor[rgb]{0.97, 0.99, 0.96} \color{black} 1.40 & 18.5\% & div & \cellcolor[rgb]{1.00, 0.93, 0.90} \color{black} -1.40 & 0.3\% \\
                    bone & \cellcolor[rgb]{0.97, 0.99, 0.96} \color{black} 1.40 & 0.2\% & pipe & \cellcolor[rgb]{1.00, 0.93, 0.90} \color{black} -1.40 & 0.4\% \\
                    goat & \cellcolor[rgb]{0.97, 0.99, 0.96} \color{black} 1.39 & 0.1\% & meters & \cellcolor[rgb]{1.00, 0.96, 0.94} \color{black} -1.37 & 0.5\% \\
                    \hline
                \end{tabular}
            }

            \vspace{1em}

            \resizebox{\columnwidth}{!}{
                \begin{tabular}{c c c ? c c c}
                    \hline
                    \multicolumn{6}{c}{\textbf{\olmo{}}} \\
                    \hline
                    \multicolumn{3}{c?}{\textbf{High Warmth}} & \multicolumn{3}{c}{\textbf{Low Warmth}} \\
                    Term & $z$ & $N$ & Term & $z$ & $N$ \\
                    \hline
                    you & \cellcolor[rgb]{0.60, 0.83, 0.58} \color{black} 2.28 & 32.0\% & italian & \cellcolor[rgb]{0.40, 0.00, 0.05} \color{white} -2.18 & 0.5\% \\
                    to & \cellcolor[rgb]{0.75, 0.90, 0.72} \color{black} 2.00 & 80.7\% & microwave & \cellcolor[rgb]{0.88, 0.18, 0.15} \color{white} -1.92 & 0.4\% \\
                    email & \cellcolor[rgb]{0.89, 0.96, 0.87} \color{black} 1.69 & 1.3\% & n1 & \cellcolor[rgb]{0.95, 0.29, 0.21} \color{white} -1.84 & 0.4\% \\
                    comma & \cellcolor[rgb]{0.90, 0.96, 0.89} \color{black} 1.64 & 1.8\% & bond & \cellcolor[rgb]{0.95, 0.29, 0.21} \color{white} -1.84 & 0.4\% \\
                    can & \cellcolor[rgb]{0.91, 0.96, 0.89} \color{black} 1.64 & 17.4\% & integrity & \cellcolor[rgb]{0.99, 0.53, 0.40} \color{black} -1.70 & 0.3\% \\
                    separated & \cellcolor[rgb]{0.91, 0.96, 0.89} \color{black} 1.63 & 2.0\% & mbr & \cellcolor[rgb]{0.99, 0.53, 0.40} \color{black} -1.70 & 0.3\% \\
                    tools & \cellcolor[rgb]{0.93, 0.97, 0.91} \color{black} 1.54 & 1.7\% & risk & \cellcolor[rgb]{0.99, 0.56, 0.44} \color{black} -1.68 & 1.2\% \\
                    needed & \cellcolor[rgb]{0.94, 0.98, 0.93} \color{black} 1.49 & 2.4\% & cups & \cellcolor[rgb]{0.99, 0.77, 0.68} \color{black} -1.54 & 0.3\% \\
                    prize & \cellcolor[rgb]{0.95, 0.98, 0.93} \color{black} 1.48 & 0.3\% & pipe & \cellcolor[rgb]{1.00, 0.86, 0.80} \color{black} -1.48 & 0.4\% \\
                    blocks & \cellcolor[rgb]{0.97, 0.99, 0.96} \color{black} 1.40 & 0.3\% & meters & \cellcolor[rgb]{1.00, 0.88, 0.83} \color{black} -1.47 & 0.5\% \\
                    \hline
                \end{tabular}
            }
            \caption{Log-odds-ratio with IDP results using tokens among warmth subsets. $z$-scores reflect the extent that each category is associated with the high (positive) or low (negative) warmth subsets, and $f$ reflects term frequency in the full corpus.}
            \label{tab:warmth-tok-lmsys}
        \end{table}

        \begin{table}[htbp]
            \centering
            \resizebox{\columnwidth}{!}{
                \begin{tabular}{c c c ? c c c}
                    \hline
                    \multicolumn{6}{c}{\textbf{\llamaBig{}}} \\
                    \hline
                    \multicolumn{3}{c?}{\textbf{High Competence}} & \multicolumn{3}{c}{\textbf{Low Competence}} \\
                    Term & $z$ & $N$ & Term & $z$ & $N$ \\
                    \hline
                    what & \cellcolor[rgb]{0.00, 0.27, 0.11} \color{white} 3.62 & 20.0\% & n1 & \cellcolor[rgb]{0.94, 0.25, 0.18} \color{white} -1.87 & 0.4\% \\
                    microwave & \cellcolor[rgb]{0.64, 0.86, 0.62} \color{black} 2.20 & 0.4\% & mbr & \cellcolor[rgb]{0.99, 0.48, 0.36} \color{black} -1.73 & 0.3\% \\
                    can & \cellcolor[rgb]{0.71, 0.88, 0.68} \color{black} 2.08 & 17.3\% & risk & \cellcolor[rgb]{0.99, 0.63, 0.51} \color{black} -1.64 & 1.2\% \\
                    bladder & \cellcolor[rgb]{0.79, 0.92, 0.76} \color{black} 1.93 & 0.4\% & write & \cellcolor[rgb]{1.00, 0.88, 0.83} \color{black} -1.46 & 17.2\% \\
                    integrity & \cellcolor[rgb]{0.80, 0.92, 0.77} \color{black} 1.91 & 0.3\% & bond & \cellcolor[rgb]{1.00, 0.90, 0.86} \color{black} -1.44 & 0.4\% \\
                    do & \cellcolor[rgb]{0.90, 0.96, 0.88} \color{black} 1.66 & 10.4\% & skirt & \cellcolor[rgb]{1.00, 0.90, 0.86} \color{black} -1.44 & 0.2\% \\
                    antenna & \cellcolor[rgb]{0.92, 0.97, 0.90} \color{black} 1.59 & 0.2\% & citations & \cellcolor[rgb]{1.00, 0.90, 0.86} \color{black} -1.44 & 0.2\% \\
                    tube & \cellcolor[rgb]{0.92, 0.97, 0.90} \color{black} 1.58 & 0.3\% & cameras & \cellcolor[rgb]{1.00, 0.95, 0.92} \color{black} -1.39 & 0.3\% \\
                    popcorn & \cellcolor[rgb]{0.93, 0.97, 0.91} \color{black} 1.56 & 0.2\% & wavelength & \cellcolor[rgb]{1.00, 0.95, 0.93} \color{black} -1.38 & 0.2\% \\
                    conda & \cellcolor[rgb]{0.93, 0.97, 0.91} \color{black} 1.56 & 0.3\% & saving & \cellcolor[rgb]{1.00, 0.95, 0.93} \color{black} -1.38 & 0.2\% \\
                    \hline
                \end{tabular}
            }

            \vspace{1em}

            \resizebox{\columnwidth}{!}{
                \begin{tabular}{c c c ? c c c}
                    \hline
                    \multicolumn{6}{c}{\textbf{\llamaSmall{}}} \\
                    \hline
                    \multicolumn{3}{c?}{\textbf{High Competence}} & \multicolumn{3}{c}{\textbf{Low Competence}} \\
                    Term & $z$ & $N$ & Term & $z$ & $N$ \\
                    \hline
                    you & \cellcolor[rgb]{0.46, 0.77, 0.47} \color{black} 2.50 & 32.0\% & microwave & \cellcolor[rgb]{0.94, 0.25, 0.19} \color{white} -1.87 & 0.4\% \\
                    the & \cellcolor[rgb]{0.65, 0.86, 0.62} \color{black} 2.19 & 122.0\% & bond & \cellcolor[rgb]{0.97, 0.36, 0.26} \color{black} -1.81 & 0.4\% \\
                    to & \cellcolor[rgb]{0.67, 0.87, 0.64} \color{black} 2.16 & 80.8\% & n1 & \cellcolor[rgb]{0.97, 0.36, 0.26} \color{black} -1.81 & 0.4\% \\
                    and & \cellcolor[rgb]{0.89, 0.96, 0.87} \color{black} 1.69 & 65.9\% & mbr & \cellcolor[rgb]{0.99, 0.58, 0.46} \color{black} -1.67 & 0.3\% \\
                    my & \cellcolor[rgb]{0.91, 0.96, 0.89} \color{black} 1.62 & 9.8\% & lights & \cellcolor[rgb]{0.99, 0.75, 0.65} \color{black} -1.56 & 0.4\% \\
                    accessory & \cellcolor[rgb]{0.91, 0.97, 0.89} \color{black} 1.62 & 0.2\% & risk & \cellcolor[rgb]{0.99, 0.85, 0.79} \color{black} -1.49 & 1.2\% \\
                    river & \cellcolor[rgb]{0.95, 0.98, 0.94} \color{black} 1.47 & 0.4\% & brand & \cellcolor[rgb]{1.00, 0.91, 0.87} \color{black} -1.43 & 1.1\% \\
                    are & \cellcolor[rgb]{0.97, 0.99, 0.96} \color{black} 1.40 & 18.5\% & div & \cellcolor[rgb]{1.00, 0.93, 0.90} \color{black} -1.40 & 0.3\% \\
                    bone & \cellcolor[rgb]{0.97, 0.99, 0.96} \color{black} 1.40 & 0.2\% & pipe & \cellcolor[rgb]{1.00, 0.93, 0.90} \color{black} -1.40 & 0.4\% \\
                    goat & \cellcolor[rgb]{0.97, 0.99, 0.96} \color{black} 1.39 & 0.1\% & meters & \cellcolor[rgb]{1.00, 0.96, 0.94} \color{black} -1.37 & 0.5\% \\
                    \hline
                \end{tabular}
            }

            \vspace{1em}

            \resizebox{\columnwidth}{!}{
                \begin{tabular}{c c c ? c c c}
                    \hline
                    \multicolumn{6}{c}{\textbf{\olmo{}}} \\
                    \hline
                    \multicolumn{3}{c?}{\textbf{High Competence}} & \multicolumn{3}{c}{\textbf{Low Competence}} \\
                    Term & $z$ & $N$ & Term & $z$ & $N$ \\
                    \hline
                    you & \cellcolor[rgb]{0.60, 0.83, 0.58} \color{black} 2.28 & 32.0\% & italian & \cellcolor[rgb]{0.40, 0.00, 0.05} \color{white} -2.18 & 0.5\% \\
                    to & \cellcolor[rgb]{0.75, 0.90, 0.72} \color{black} 2.00 & 80.7\% & microwave & \cellcolor[rgb]{0.88, 0.18, 0.15} \color{white} -1.92 & 0.4\% \\
                    email & \cellcolor[rgb]{0.89, 0.96, 0.87} \color{black} 1.69 & 1.3\% & n1 & \cellcolor[rgb]{0.95, 0.29, 0.21} \color{white} -1.84 & 0.4\% \\
                    comma & \cellcolor[rgb]{0.90, 0.96, 0.89} \color{black} 1.64 & 1.8\% & bond & \cellcolor[rgb]{0.95, 0.29, 0.21} \color{white} -1.84 & 0.4\% \\
                    can & \cellcolor[rgb]{0.91, 0.96, 0.89} \color{black} 1.64 & 17.4\% & integrity & \cellcolor[rgb]{0.99, 0.53, 0.40} \color{black} -1.70 & 0.3\% \\
                    separated & \cellcolor[rgb]{0.91, 0.96, 0.89} \color{black} 1.63 & 2.0\% & mbr & \cellcolor[rgb]{0.99, 0.53, 0.40} \color{black} -1.70 & 0.3\% \\
                    tools & \cellcolor[rgb]{0.93, 0.97, 0.91} \color{black} 1.54 & 1.7\% & risk & \cellcolor[rgb]{0.99, 0.56, 0.44} \color{black} -1.68 & 1.2\% \\
                    needed & \cellcolor[rgb]{0.94, 0.98, 0.93} \color{black} 1.49 & 2.4\% & cups & \cellcolor[rgb]{0.99, 0.77, 0.68} \color{black} -1.54 & 0.3\% \\
                    prize & \cellcolor[rgb]{0.95, 0.98, 0.93} \color{black} 1.48 & 0.3\% & pipe & \cellcolor[rgb]{1.00, 0.86, 0.80} \color{black} -1.48 & 0.4\% \\
                    blocks & \cellcolor[rgb]{0.97, 0.99, 0.96} \color{black} 1.40 & 0.3\% & meters & \cellcolor[rgb]{1.00, 0.88, 0.83} \color{black} -1.47 & 0.5\% \\
                    \hline
                \end{tabular}
            }
            \caption{Log-odds-ratio with IDP results using tokens among competence subsets. $z$-scores reflect the extent that each category is associated with the high (positive) or low (negative) warmth subsets, and $f$ reflects term frequency in the full corpus.}
            \label{tab:comp-tok-lmsys}
        \end{table}